\pdfoutput=1

\documentclass[11pt]{article}

\usepackage[final]{acl}

\usepackage{times}
\usepackage{latexsym}

\usepackage[T1]{fontenc}

\usepackage[utf8]{inputenc}

\usepackage{microtype}

\usepackage{inconsolata}

\usepackage{graphicx}
\usepackage{subcaption}
\usepackage{geometry}
\usepackage{amsmath}
\usepackage{amssymb}
\usepackage{tabularx}
\usepackage{colortbl}
\usepackage{xcolor}
\usepackage{array}
\usepackage{algorithm}
\usepackage{algpseudocode}
\usepackage{enumitem}
\usepackage{verbatim}
\usepackage{pgfplots}
\usepackage{tikz}
\pgfplotsset{compat=1.18}
\title{ECHO-LLaMA: Efficient Caching for High-Performance LLaMA Training}

\author{
  \textbf{Maryam Dialameh}\textsuperscript{1,2}, 
  \textbf{Rezaul Karim}\textsuperscript{2}, 
  \textbf{Hossein Rajabzadeh}\textsuperscript{1,2}, 
  \textbf{Omar Mohamed Awad}\textsuperscript{2},\\
  \textbf{Hyock Ju Kwon}\textsuperscript{1},  
  \textbf{Boxing Chen}\textsuperscript{2}, 
  \textbf{Walid Ahmed}\textsuperscript{2},  
  \textbf{Yang Liu}\textsuperscript{2} \\
  \textsuperscript{1}University of Waterloo, Waterloo, Canada \\
  \textsuperscript{2}Ascend Team, Huawei Technologies, Toronto, Canada \\
  \texttt{\{maryam.dialameh, hossein.rajabzadeh, hjkwon\}@uwaterloo.ca} \\
  \texttt{\{rezaul.karim3, walid.ahmed1, boxing.chen, yang.liu8\}@huawei.com}
}

\begin{document}
\maketitle
\begin{abstract}
This paper introduces ECHO-LLaMA, an efficient LLaMA architecture designed to improve  both the training speed and inference throughput of LLaMA architectures while maintaining its learning capacity. ECHO-LLaMA transforms LLaMA models into shared KV caching across certain layers, significantly reducing KV computational complexity while maintaining or improving language performance. Experimental results demonstrate that ECHO-LLaMA achieves up to 77\% higher token-per-second throughput during training, up to 16\% higher Model FLOPs Utilization (MFU), and up to 14\% lower loss when trained on an equal number of tokens. Furthermore, on the 1.1B model, ECHO-LLaMA delivers approximately 7\% higher test-time throughput compared to the baseline. By introducing a computationally efficient adaptation mechanism, ECHO-LLaMA offers a scalable and cost-effective solution for pretraining and finetuning large language models, enabling faster and more resource-efficient training without compromising performance.
\end{abstract}

 \begin{figure}[ht]
    \centering
    \includegraphics[width=0.99\columnwidth, height=0.99\columnwidth]{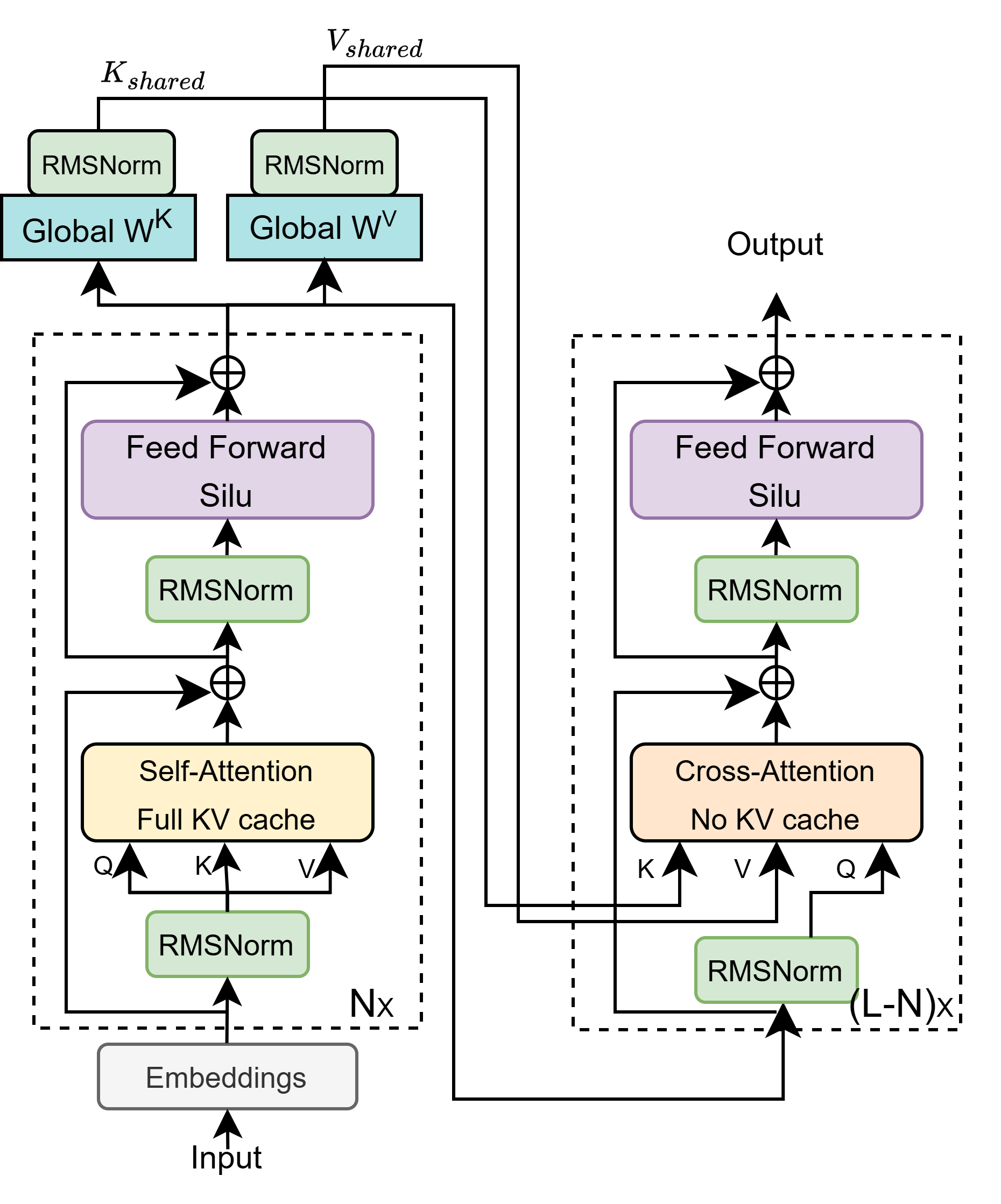}
    \caption{ECHO-LLaMA: The architecture uses shared KV caching coupled with layer-wise adaptation, gradually converting pretrained LLaMA models into ECHO-LLaMA architectures. The number of shared KV layers are determined by the total number of layers ($L$) minus the number of self-attention layers ($N$). ECHO-LLaMA reduces computational overhead while improving the inference speed. 
    }
    \label{fig:yoco-llama-workflow}
\end{figure}

\section{Introduction}

Large language models (LLMs) have shown remarkable success across a wide range of natural language processing (NLP) tasks, including text generation \cite{makridakis2023large}, summarization \cite{zhang2024benchmarking}, and question answering and more \cite{chang2024survey}. Despite their capabilities, training and deploying these models is highly resource-intensive, demanding significant computational power and memory \cite{chowdhery2023palm}. For instance, a transformers-based LLM requires about three terabyte KV-cached memory for a model of size 500B with 8k context length and 128 batch size \cite{pope2023efficiently}. Furthermore, the process of pretraining for LLMs is both resource and data intensive \cite{milano2023large,hoffmann2022training}. 

Several existing works have been proposed to address the challenge of improving transformer efficiency \cite{zhang2023h2o,yang2023gated,lee2024infinigen,tang2024razorattention,adnan2024keyformer,hajimolahoseini2023training,ahmed2023speeding}, including GQKVA \cite{javadi2023gqkva}, Beyond KV Caching \cite{liao2024beyond} and EchoAtt \cite{rajabzadeh2024echoatt} that specifically share the attention weights across layers to decrease computation and parameters; however, they still need to cache for V matrices. YOCO (You Only Cache Once) is another similar work that investigates sharing key-value (KV) caches across the second-half of layers (cross-decoder layers) while employing windowed-attention for the first half; thereby gaining computational and memory efficiency during training and inference~\cite{sun2024you}.
 YOCO models, however, use windowed self-attention, which limits their ability to capture long-range token dependencies \cite{beltagy2020longformer}. Furthermore, they apply shared KV caches to exactly half of the layers, enforcing a rigid and suboptimal KV sharing strategy. This lack of flexibility is particularly problematic for edge devices, which may benefit from adaptive KV sharing, allowing for shorter or longer KV reuse based on resource constraints and workload demands.
Inspired by YOCO, this paper introduces ECHO-LLaMA, an efficient modification of LLaMA architectures that focuses on having efficient  training and pretraining processes of LLaMA models. ECHO-LLaMA leverages a caching strategy where KV representations are shared across cross-decoder layers (typically the second half of the LLaMA layers), effectively reducing memory overhead and computational redundancy. Unlike the original YOCO approach \cite{sun2024you}, which is designed for training models from scratch, ECHO-LLaMA extends this methodology to pre-trained LLaMA models, enabling a more efficient adaptation without requiring full re-training. This is achieved through incremental adaptation, a strategy that incrementally transitions pre-trained LLaMA models into the ECHO architecture while maintaining or improving their language capabilities. The results demonstrate significant improvements in training throughput (up to 77\%), Model FLOPs Utilization (MFU, up to 16\%), and lower loss (up to 14\%) compared to baselines. Moreover, ECHO-LLaMA achieves faster test-time throughput (approximately 7\% on the 1.1B model), further showcasing its effectiveness for large-scale model deployment. By reducing training costs, accelerating inference, and preserving model performance, ECHO-LLaMA represents a scalable and practical solution for optimizing pre-trained large language models.

The proposed ECHO-LLaMA framework offers several significant advantages. First, it achieves faster inference without sacrificing language performance. Second, it reduces the need for heavy pre-training, making it a practical approach for building efficient versions of LLaMA models. Third, experiments conducted on "Nvidia-V100 GPUs" and "Huawei Ascend NPU-910B" devices show that ECHO-LLaMA consistently achieves higher training throughput and competitive performance compared to LLaMA baselines. Additionally, the results demonstrate the ability of the proposed framework to maintain or improve model performance while reducing computational costs. Hence, the main contributions of this work are summarized as follows:
\begin{itemize}
  \item We propose ECHO-LLaMA, an efficient LLaMA architecture that shares KV caches across selected set of layers to reduce computational redundancy.
  \item ECHO-LLaMA employs a layer-wise incremental adaptation for the training strategy to efficiently convert pre-trained LLaMA models into the ECHO-LLaMA structure.
  \item This approach significantly improves training throughput and inference speed without compromising language performance.
  \item Extensive experiments demonstrate higher model FLOPs utilization, lower loss, and increased tokens-per-second throughput compared to baseline models.
\end{itemize}

\noindent This paper is organized as follows. Section~\ref{sec:method} details the proposed ECHO-LLaMA architecture and its incremental adaptation strategy for converting pretrained models into efficient structures. In Section~\ref{sec:experiments}, we present extensive experimental evaluations demonstrating improvements in training throughput, model FLOPs utilization, and language performance. Sections \ref{sec:discussion} and \ref{sec:limitations} discuss  our proposed method and its potential limitations. Finally,  Section~\ref{sec:conclusion} concludes the paper with insights and directions for future research.

\section{Proposed Method}
\label{sec:method}
 This section introduces our efficient framework for converting pretrained LLaMA models into ECHO-LLaMA architectures through incremental layer-wise adaption. The core idea consists of two parts: 1) the ECHO-LLaMA architecture and 2) layer-wise incremental adaptation strategy. Figure~\ref{fig:yoco-llama-workflow} illustrates the workflow of the ECHO-LLaMA, in which the first half of the layers remains full self-attention with full KV caches, and the second half of layers are gradually converted into cross-decoders with only one set of global KV caches. Each cross-decoder layer uses the global KV, computing by linearly transforming the output of the middle layer followed by an RMS normalization \cite{zhang2019root}.

Given an input sequence of token embeddings $x_1, x_2, \dots, x_n \in R^{d}$, where $d$ is the hidden size, the first $N$ layers of an ECHO-LLaMA model (out of a total of $L$ layers) operate similarly to standard LLaMA layers. The parameter $N \in \{L/2, \dots, L\}$ is a hyperparameter that controls the extent of KV sharing in ECHO-LLaMA. Let's $X_{l-1}$ be the output of layer $l-1$, which would be the input to layer $l$ Therefore, for an input $X_{l-1}$, $l\in\{0,1,...,N\}$, we have:

\begin{equation}
\resizebox{\columnwidth}{!}{$
\left\{
\begin{array}{rcl}
X'_l &=& \text{Self-Attn}\bigl(\text{RMSNorm}(X_{l-1})\bigr) + X_{l-1}, \\[1ex]
X_l  &=& \text{SFF}\bigl(\text{RMSNorm}(X'_l)\bigr) + X'_l.
\end{array}
\right.
$}
\label{eq:transformer_block}
\end{equation}

\vspace{1em}

\begin{equation}
\resizebox{\columnwidth}{!}{$
\text{Self-Attn}(X) = \text{softmax}\!\left(\frac{XW_Q \, (XW_K)^T}{\sqrt{d_k}}\right) XW_V.
$}
\label{eq:self_attention}
\end{equation}


\noindent where $W_Q, W_K, W_V \in R^{d_{\text{model}} \times d_k}$, $d_k$ represents the dimensionality of the key and query vectors, typically set as $d_k = \frac{d_{\text{model}}}{h}$ and $h$ is the number of attention heads. The LLaMA MLP module is a SiLU \cite{elfwing2018sigmoid} activated feed forward module, SFF(X), consisting of a gate projection, an up projection, and a down projection. Given an input tensor \( X \), the MLP transformation is defined as:
\begin{equation}
\resizebox{.85\columnwidth}{!}{$
    \begin{aligned}
        SFF(X) &= W_{\text{down}} \Big( SiLU(W_{\text{gate}} X + b_{\text{gate}}) \odot \\
               &\quad (W_{\text{up}} X + b_{\text{up}}) \Big) + b_{\text{down}}
    \end{aligned}
$}
\end{equation}

\noindent where \( W_{\text{gate}} \in \mathbb{R}^{d \times d_{\text{up}}} \), \( b_{\text{gate}} \in \mathbb{R}^{d_{\text{up}}} \) (bias for gate projection parameters), \( W_{\text{up}} \in \mathbb{R}^{d \times d_{\text{up}}} \), \( b_{\text{up}} \in \mathbb{R}^{d_{\text{up}}} \) (up projection parameters), \( W_{\text{down}} \in \mathbb{R}^{d_{\text{up}} \times d} \), \( b_{\text{down}} \in \mathbb{R}^{d} \) (down projection parameters), and \( \odot \) represents the element-wise Hadamard product.


\noindent The output of layer-$N$, i.e. $X_N$, is then passed through global $W^K$ and $W^V$ followed by RMS normalization, creating one shared KV for the rest of layers. In other words, layers $l\in\{N+1,...,L\}$ use the same KV matrices and, therefore, compute cross attention between query and shared KV. This modification allows LLaMA models to save KV cache memory and increase both training and inference speed. Assuming $X_N$ as the output of layer $N$, the shared KV in ECHO-LLaMA is computed as follows:

\begin{align}
    K_{\text{shared}} &= \text{RMSNorm}( W^K_{\text{global}}X_N) \\
    V_{\text{shared}} &= \text{RMSNorm}( W^V_{\text{global}}X_N)
\end{align}

\noindent where $W^K_{\text{global}}$ and $W^V_{\text{global}}$ follows the same size as $W^K$ and $W^V$. We further modify the LLaMA architecture to compute cross attention between query and share KV as follows:

\begin{equation}
\resizebox{\columnwidth}{!}{$
\text{Cross-Attn}(X) = \text{softmax}\!\left(\frac{X_NW_Q \, (K_{shared})^T}{\sqrt{d_k}}\right) V_{shared}.
$}
\label{eq:self_attention}
\end{equation}

\noindent As depicted in Figure \ref{fig:yoco-llama-workflow}, the output of cross attention is then added by a residual and passed through RMS normalization and $SFF$ module.

\begin{algorithm}[t]
\caption{Incremental Adaptation for ECHO-LLaMA}
\label{alg:incremental_adaptation}
\begin{algorithmic}[1]
\Require Pretrained model $\mathcal{F}_{\theta_{\mathrm{pre}}}$, total layers $L$, threshold layer $N$ for cross-decoder conversion, token budget per stage $\mathcal{T}_{\mathrm{stage}} \approx 4\text{M}$, training steps per stage $S \approx 150$, final token budget $\mathcal{T}_{\mathrm{final}} \approx 4\text{B}$
\Ensure Adapted model $\mathcal{F}_\theta$
\State Initialize: $\mathcal{F}_\theta \gets \mathcal{F}_{\theta_{\mathrm{pre}}}$
\State Define adaptation range: $\mathcal{R} \gets \{\ell \mid \ell = L, L-1, \dots, N\}$
\ForAll{$\ell \in \mathcal{R}$ (in descending order)}
    \State Convert layer $\ell$ into a cross-decoder layer in $\mathcal{F}_\theta$
    \State Update parameters $\theta^{(\ell)}$ by minimizing 
    \[
    \min_{\theta^{(\ell)}} \; \mathcal{L}\Bigl(\mathcal{F}_\theta, \mathcal{D}_{\ell}\Bigr)
    \]
    for $S$ steps on a token set of size $\mathcal{T}_{\mathrm{stage}}$
\EndFor
\State Fine-tune $\mathcal{F}_\theta$ for 1 epoch on a token set of size $\mathcal{T}_{\mathrm{final}}$
\State \Return $\mathcal{F}_\theta$
\end{algorithmic}
\end{algorithm}


\begin{table*}[t]
\centering
\renewcommand{\arraystretch}{1.2}
\resizebox{\textwidth}{!}{%
\begin{tabular}{|l|c|c|c|c|c|c|c|c|c|c|}
\hline
\rowcolor{green!50} 
\textbf{Model} & \textbf{CQA} & \textbf{BQ} & \textbf{WG} & \textbf{PiQA} & \textbf{Arc\_c} & \textbf{Arc\_e} & \textbf{OBQA} & \textbf{HS} & \textbf{Avg. Acc}& \textbf{Throughput} \\ \hline
\multicolumn{11}{|c|}{\cellcolor{gray!30} \textbf{Zero-Shot Results}} \\ \hline
TinyLLaMA (Baseline) & 20.15 & 56.02 & 59.35 & 72.63 & 32.68 & \textbf{55.47} & \textbf{36.80} & \textbf{61.47} & 49.32 &NA\\ \hline
ECHO-TinyLLaMA-25\%-Shared-KV & \textbf{20.72} & \textbf{58.86} & \textbf{59.75} & \textbf{73.45} & \textbf{33.19} & 54.12 & 36.40 & 59.01 & \textbf{49.44} &\textbf{3.35\%} \textcolor{green}{\ensuremath{\uparrow}}\\ \hline
ECHO-TinyLLaMA-50\%-Shared-KV & \textbf{20.64} & \textbf{58.69} & 58.96 & \textbf{73.72} & 31.74 & 53.54 & 35.00 & 59.01 & 48.91 & \textbf{7\%} \textcolor{green}{\ensuremath{\uparrow}}\\ \hline
\multicolumn{11}{|c|}{\cellcolor{gray!30} \textbf{5-Shot Results}} \\ \hline
TinyLLaMA1.1B-V1.1 (Baseline) & 19.0 & 62.11 & 62.12 & 74.65 & 36.43 & \textbf{69.23} & \textbf{38.20} & \textbf{62.19} & \textbf{52.98} &NA\\ \hline
ECHO-TinyLLaMA-25\%-Shared-KV & \textbf{19.10} & \textbf{63.41} & \textbf{62.16} & \textbf{74.67} & \textbf{36.50} & 68.14 & 36.89 & 59.35 & 52.53 &\textbf{3.35}\%\textcolor{green}{\ensuremath{\uparrow}}\\ \hline
ECHO-TinyLLaMA-50\%-Shared-KV & 18.10 & \textbf{62.39} & 61.48 & 73.78 & 36.35 & 66.20 & 37.00 & 59.93 & 51.79 &\textbf{7\%}\textcolor{green}{\ensuremath{\uparrow}}\\ \hline
\end{tabular}%
}
\caption{Zero-shot and 5-shot evaluation results for different models across multiple benchmarks. The last column reports the test throughput improvement compared to the baseline in terms of generated tokens per seconds. NA means not applicable. Additional evaluation results on larger models and more diverse datasets are included in Appendix~\ref{more-exp}.}
\label{tab:zero_shot_results}
\end{table*}

\subsection{Incremental Adaptation Strategy}
ECHO-LLaMA leverages incremental adaptation from a pre-trained LLaMA model. Denote the parameters of layer \(\ell\) by \(\theta^{(\ell)}\). The layers \(\{1,\dots,L\}\) are partitioned into ordered blocks \(\{\mathcal{B}_1, \mathcal{B}_2, \dots\}\), where each block \(\mathcal{B}_m\) is adapted in a separate stage. In our experiments, we assume the size of $B_m$ is 1. The incremental adaptation process is defined as follows:
\begin{enumerate}[label=\arabic*.]
    \item \textbf{Initialization}: For all \(\ell\), set \(\theta^{(\ell)} \leftarrow \theta^{(\ell)}_{\text{pretrained}}\).
    \item \textbf{Stage \(m\) Update}: For each \(\ell \in \mathcal{B}_m\), update \(\theta^{(\ell)}\) via gradient descent on a data subset \(\mathcal{D}_m \subseteq \mathcal{D}\), while keeping parameters in layers \(\ell \notin \mathcal{B}_m\) frozen.
\end{enumerate}
The stage-\(m\) objective is:
\begin{equation}
\min_{\{\theta^{(\ell)}: \ell \in \mathcal{B}_m\}} \; \mathcal{L}\Bigl(\mathcal{F}_\theta, \mathcal{D}_m\Bigr).
\end{equation}

The overall training objective is defined as the cross-entropy loss:
\begin{equation}
\mathcal{L}(\theta; \mathcal{D}) = \sum_{(x_i,y_i)\in \mathcal{D}} \sum_{t=1}^{T_i} -\log P_\theta(y_{i,t} \mid x_i, y_{i,1:t-1}),
\end{equation}
where \(T_i\) is the target sequence length for the \(i\)th sample. During incremental adaptation, only the parameters in the current block \(\mathcal{B}_m\) are updated.
Algorithm~\ref{alg:incremental_adaptation} outlines the incremental adaptation strategy used to transform a pre-trained LLaMA model into the ECHO-LLaMA architecture. Starting with the pretrained model \(\mathcal{F}_{\theta_{\mathrm{pre}}}\), we define an adaptation range 
\[
\mathcal{R} = \{ \ell \mid \ell = L, L-1, \dots, N \},
\]
where \(L\) is the total number of layers and \(N\) is the target layer at which adaptation stops. In general, each layer \(\ell \in \mathcal{R}\) can be grouped as a block \(\mathcal{B}_m\) in a block-wise incremental adaptation strategy. For each layer (or block \(\mathcal{B}_m\)) in \(\mathcal{R}\), the standard self-attention is converted into cross-attention by introducing global shared KV matrices. The parameters \(\theta^{(\ell)}\) of the converted layer are then updated by minimizing the loss function 
\[
\mathcal{L}\Bigl(\mathcal{F}_\theta, \mathcal{D}_{\ell}\Bigr)
\]
for \(S \approx 150\) steps using a token budget of \(\mathcal{T}_{\mathrm{stage}} \approx 4\text{M}\) tokens. This incremental update enables the model to effectively adapt to the new cross-attention mechanism while mitigating catastrophic forgetting \cite{kirkpatrick2017overcoming}.

\noindent Once all layers in \(\mathcal{R}\) have been incrementally adapted, the entire model \(\mathcal{F}_\theta\) undergoes a final fine-tuning phase for one epoch on a larger token set (\(\mathcal{T}_{\mathrm{final}} \approx 4\text{B}\) tokens). This final phase allows the model to further refine its representations, stabilize training dynamics, and enhance overall generalization for downstream tasks. By incrementally updating layers (or blocks) with approximately 150 training steps per stage, the method achieves a favorable balance between computational cost and performance, preserving critical information flow and ensuring robust cross-attention mechanisms.

\subsubsection{Memory Footprint}
In a standard LLaMA model, each layer \(\ell\) has its own key and value projection matrices, incurring a memory cost proportional to:
\[
L \times (2d^2).
\]
For ECHO-LLaMA, with partial KV sharing, the cost becomes:
\[
(1-p)L \times (2d^2) + 2d^2.
\]
Thus, the ratio of KV memory usage in ECHO-LLaMA versus the baseline is:
\[
\frac{(1-p)L \times (2d^2) + 2d^2}{L \times (2d^2)} = (1-p) + \frac{1}{L}.
\]
For large \(L\), this ratio approximates \(1-p\).

\section{Experiments}
\label{sec:experiments}
\textbf{Benchmarks-} The evaluation benchmarks used to assess the performance of ECHO-TinyLLaMA \footnote{Converting from TinyLlaMA-1.1B \url{https://huggingface.co/TinyLlama/TinyLlama_v1.1}} span a wide range of natural language understanding tasks, ensuring comprehensive coverage of different linguistic and reasoning challenges: CommonsenseQA (CQA), BoolQ (BQ), Winogrande (WG), PiQA, ARC\_c and Arc\_e (challenge and easy), OpenBookQA  (OBQA), and HellaSwag (HS). These benchmarks collectively test the model's strengths in commonsense reasoning, linguistic understanding, and scientific knowledge application. 

\textbf{TinyLLaMA MFU-} The MFU for the original TinyLLaMA and ECHO-TinyLLaMA was evaluated on different configuration of Nvidia-V100 GPUs and Ascend-910B NPUs, different batch sizes, fixed sequence length of 2048, and 100 training steps. We use the MFU calculation script from LLaMA-Factory repository \cite{zheng2024llamafactory} \footnote{Please see cal\_mfu.py in their repository.}. However, in case of the device types (GPU or NPU), this script needs several modification before computing MFU, including setting precision to fp16, increasing num\_worker for data pre-processing, setting the finetuning\_type to full, and updating the theoretical FLOPs based on your computing devices.

\noindent \textbf{Language Model Evaluations-} We used LM-Harness repository as an evaluation tool to assess the language performance of different ECHO-LLaMA architecture \cite{eval-harness}. \footnote{We acknowledge that a direct comparison with YOCO models would be valuable; however, no official checkpoints have been released publicly, preventing such an evaluation.
A direct comparison with YOCO models was not possible, as no official checkpoints for YOCO have been released publicly.}

To evaluate the effectiveness of our incremental  training strategy, we incrementally applied the training on approximately 4 billion tokens \footnote{Taken from \url{https://huggingface.co/datasets/cerebras/SlimPajama-627B/tree/main/train}} , initializing the ECHO-TinyLLaMA models from the baseline weights \cite{huggingfaceTinyLlama_v11}. We experiment for flexibly growing KV sharing architectures allowing the cross-decoders to be used for 25\% and 50\% of the layers in contrast to a fixed sharing size of YOCO. 
For 25\% configuration, only the last 25\% of layers are converted into cross-decoder layers. In the case of the 50\% configuration, the entire second half of the layers are transitioned into cross-decoders. 

The evaluation results, presented in Table~\ref{tab:zero_shot_results}, demonstrate that \textbf{ECHO-TinyLLaMA with 25\% shared cross-decoders consistently outperforms the baseline} in both zero-shot and 5-shot settings, achieving the highest average accuracy in the zero-shot case and performing competitively in the 5-shot scenario. Notably, \textbf{ECHO-TinyLLaMA with 50\% cross-decoders} closely matches the baseline performance in both evaluation setups, with an average accuracy gap of less than 0.5\% in the zero-shot setting and about 1.2\% in the 5-shot case. Despite this negligible drop, the 50\% configuration delivers a \textbf{7\% throughput improvement}, highlighting its effectiveness in balancing generation speed with competitive accuracy.

\noindent Appendix \ref{more-exp} includes more results on larger models and more diverse datasets. Moreover, Appendix~\ref{appendix:FS-vs-Inc} provides a comparison results showing that the incremental strategy outperforms the full-stage approach, where all designated layers are converted to shared-KV at once.

\begin{table*}[t]
\centering
\footnotesize 
\renewcommand{\arraystretch}{1.2} 
\renewcommand{\tabcolsep}{4pt} 
\begin{tabular}{|l|c|c|c|c|}
\hline
\rowcolor{orange!80} \textbf{Model}  & \textbf{Train Loss} & \multicolumn{3}{c|}{\textbf{MFU (\%)}} \\ \cline{3-5}
\rowcolor{orange!80}                &                     & \textbf{1 NPU} & \textbf{4 NPUs} & \textbf{8 NPUs} \\ \hline
LLaMA-125M (baseline)      & \textbf{5.25} & 14.35\newline {\tiny Bs=58} & 9.19\newline {\tiny Bs=50} & 8.33\newline {\tiny Bs=50} \\ \hline
ECHO-LLaMA-125M           & 5.28          & \textbf{14.49}\newline {\tiny Bs=58} & \textbf{9.62}\newline {\tiny Bs=50} & \textbf{9.45}\newline {\tiny Bs=55} \\ \hline
TinyLLaMA (baseline)      & 4.65          & 24.17\newline {\tiny Bs=20} & 22.11\newline {\tiny \textbf{Bs=15}} & 22.19\newline {\tiny Bs=40} \\ \hline
ECHO-TinyLLaMA            & \textbf{4.45} & \textbf{27.10}\newline {\tiny Bs=36} & \textbf{27.98}\newline {\tiny \textbf{Bs=20}} & \textbf{22.51}\newline {\tiny Bs=40} \\ \hline
LLaMA-3B (baseline)       & 4.25          & 29.36\newline {\tiny Bs=10} & 32.21\newline {\tiny Bs=20} & 30.89\newline {\tiny Bs=20} \\ \hline
ECHO-LLaMA-3B             & \textbf{4.15} & \textbf{30.78}\newline {\tiny Bs=12} & \textbf{48.31}\newline {\tiny Bs=30} & \textbf{46.34}\newline {\tiny Bs=30} \\ \hline
LLaMA-7B (baseline)       & 3.90          & OOM & 38.78\newline {\tiny Bs=5} & 35.07\newline {\tiny Bs=10} \\ \hline
ECHO-LLaMA-7B             & \textbf{3.34} & 35.58\newline {\tiny Bs=4} & Loss Scale Error & \textbf{35.53}\newline {\tiny Bs=18} \\ \hline
\end{tabular}
\caption{Results on NPU-910B: Comparison of ECHO and Non-ECHO LLaMA models across different model sizes and device configurations. All experiments were conducted with a sequence length of 2048. Batch sizes (Bs) are annotated in smaller font. Final training loss values are reported under an equal training budget.}
\label{tab:npu_results}
\end{table*}

\begin{table*}[t]
\centering
\footnotesize 
\renewcommand{\arraystretch}{1.2} 
\renewcommand{\tabcolsep}{4pt} 
\begin{tabular}{|l|c|c|c|c|}
\hline
\rowcolor{orange!80} \textbf{Model} & \textbf{Train Loss} & \multicolumn{3}{c|}{\textbf{MFU (\%) SDPA-Attention}} \\ \cline{3-5}
\rowcolor{orange!80}               &                     & \textbf{1 GPU} & \textbf{4 GPUs} & \textbf{8 GPUs} \\ \hline
LLaMA-125M (baseline)              & \textbf{5.06}       & 23.59          & 22.30           & 21.96           \\ \hline
ECHO-LLaMA-125M                    & 5.08                & \textbf{24.32} & \textbf{22.73}  & \textbf{22.25}  \\ \hline
TinyLLaMA (baseline)              & 4.97                & 34.34          & 34.33           & Loss Scale Error \\ \hline
ECHO-TinyLLaMA                     & \textbf{4.67}       & \textbf{39.32} & \textbf{34.42}  & Loss Scale Error \\ \hline
LLaMA-3B (baseline)               & \textbf{4.68}       & OOM            & 35.31           & 34.01           \\ \hline
ECHO-LLaMA-3B                      & 4.70                & OOM            & \textbf{37.34}  & \textbf{36.05}  \\ \hline
LLaMA-7B (baseline)               & 4.56                & OOM            & OOM             & 31.67           \\ \hline
ECHO-LLaMA-7B                      & \textbf{4.00}       & OOM            & OOM             & \textbf{35.74}  \\ \hline
\end{tabular}
\caption{Results on V100 with a sequence length of 2048 under different configurations with varying numbers of devices (1, 4, and 8). Batch size (BS) and out-of-memory (OOM) errors are reported. ECHO versions demonstrate higher MFU compared to baselines. Final training loss values are reported under an equal training budget.}
\label{tab:gpu_results}
\end{table*}

\noindent \textbf{Efficiency Comparison-}Tables \ref{tab:npu_results} and \ref{tab:gpu_results} provide a detailed comparison between ECHO and non-ECHO versions of LLaMA models of four sizes of parameters: 125M, 1.1B, 3B, and 7B. The experiments were conducted on two both Ascend-910B and Nvidia V100, each using up to 8 devices. The comparison metrics include training loss at equal train steps and training MFU. From the results, we observe the following trends:
\begin{itemize}
    \item \textbf{NPU-910B}: The ECHO-LlaMA models consistently outperform the non-ECHO counterparts across all LLM sizes. ECHO-LLaMA models achieve lower loss values and up to 15\% improvements in MFU, as evidenced by reductions in \textit{Train Speed sec/step} and increases in \textit{Train Token/sec}. For example, ECHO-LLaMA-3B improves the MFU by ~15\%.
    \item \textbf{GPU-V100}: The ECHO versions are comparable to or better than the non-ECHO counterparts in training loss. Additionally, ECHO-LLaMA models generally achieve higher training MFU, up to ~4\%. 
\end{itemize}

\begin{figure*}[t]
    \centering
\includegraphics[width=0.9\textwidth]{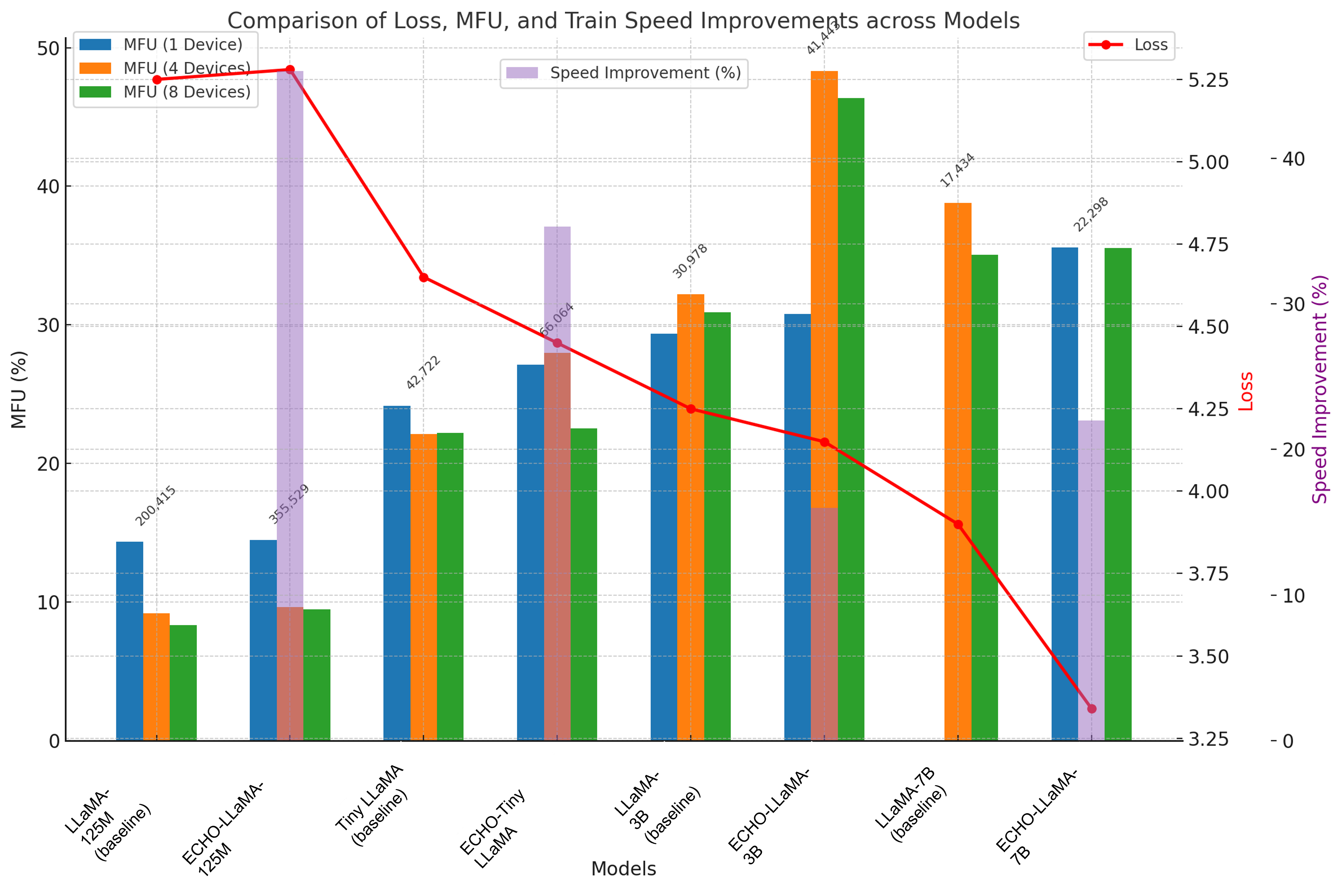}
    \caption{Comparison between of training throughput (Tokens/sec) and final loss for ECHO-LLaMA models and their baselines. ECHO versions consistently achieve lower loss with higher or comparable Tokens/sec speed. Each model is pretrained from scratch on 4B tokens through 1000 steps. Values on top each set of bars shows train token/second throughput.}
    \label{fig:npu_tokens_loss_2}
\end{figure*}

\noindent Figure \ref{fig:npu_tokens_loss_2} visually compares the training throughput (Tokens/sec), MFU, and final loss across various LLaMA models and their ECHO-LLaMA counterparts. The ECHO-LLaMA versions consistently demonstrate improvements in MFU for configurations with 1, 4, and 8 devices, while achieving lower or comparable final loss compared to their baselines. Notably, ECHO-LLaMA models, such as ECHO-LLaMA-125M and ECHO-TinyLLaMA, exhibit significant speed improvements, as indicated by the purple bars, while maintaining competitive or better loss values. These results explains the effectiveness of ECHO-LLaMA's architecture in enhancing both training efficiency and model performance.

\noindent \textbf{Scaling Law Validation-} To validate the scalability of ECHO-LLaMA models, we conducted experiments across various model sizes (125M, 1.1B, 3B, and 7B) and compared the training losses with their baseline counterparts. Each model is pretrained on 4B tokens through 1000 training steps.

\begin{figure}[!h]

\begin{tikzpicture}[scale=0.90]
    \begin{axis}[
        grid = both,
        ylabel = {\small{Train Loss}},
        xlabel = {\small{Model Size}},
        symbolic x coords={125M, 1.1B,3B,7B},
        xtick = data,
        legend cell align=left,
        legend style={
                at={(0.78,1.05)},
                legend columns=2,
                anchor=south east,
                column sep=1ex
        }
    ]
        \addplot 
        coordinates {(125M, 5.25)(1.1B, 4.65)(3B,4.25)(7B,3.90)};
        \addplot 
        coordinates {(125M, 5.28) (1.1B, 4.45)(3B,4.15)(7B,3.34)};
        \legend{\small{LLaMA}, \small{ECHO-LLaMA}}
    \end{axis}

\end{tikzpicture}
\caption{Scaling law diagram comparing training loss for ECHO-LLaMA of different sizes against the baselines.}
\label{fig:scaling_law}
\end{figure}
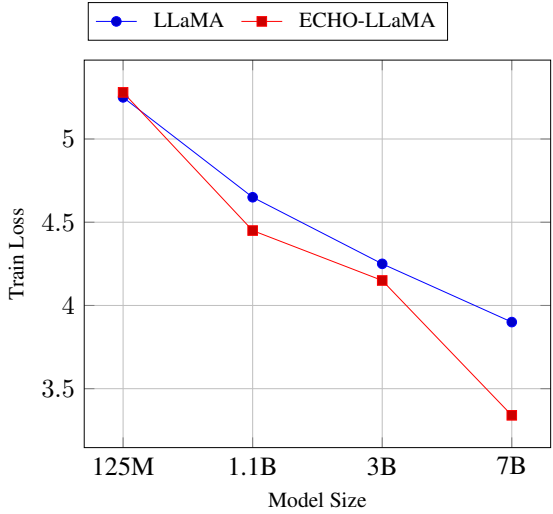

\begin{figure}[t]
    \centering
\includegraphics[width=\columnwidth]{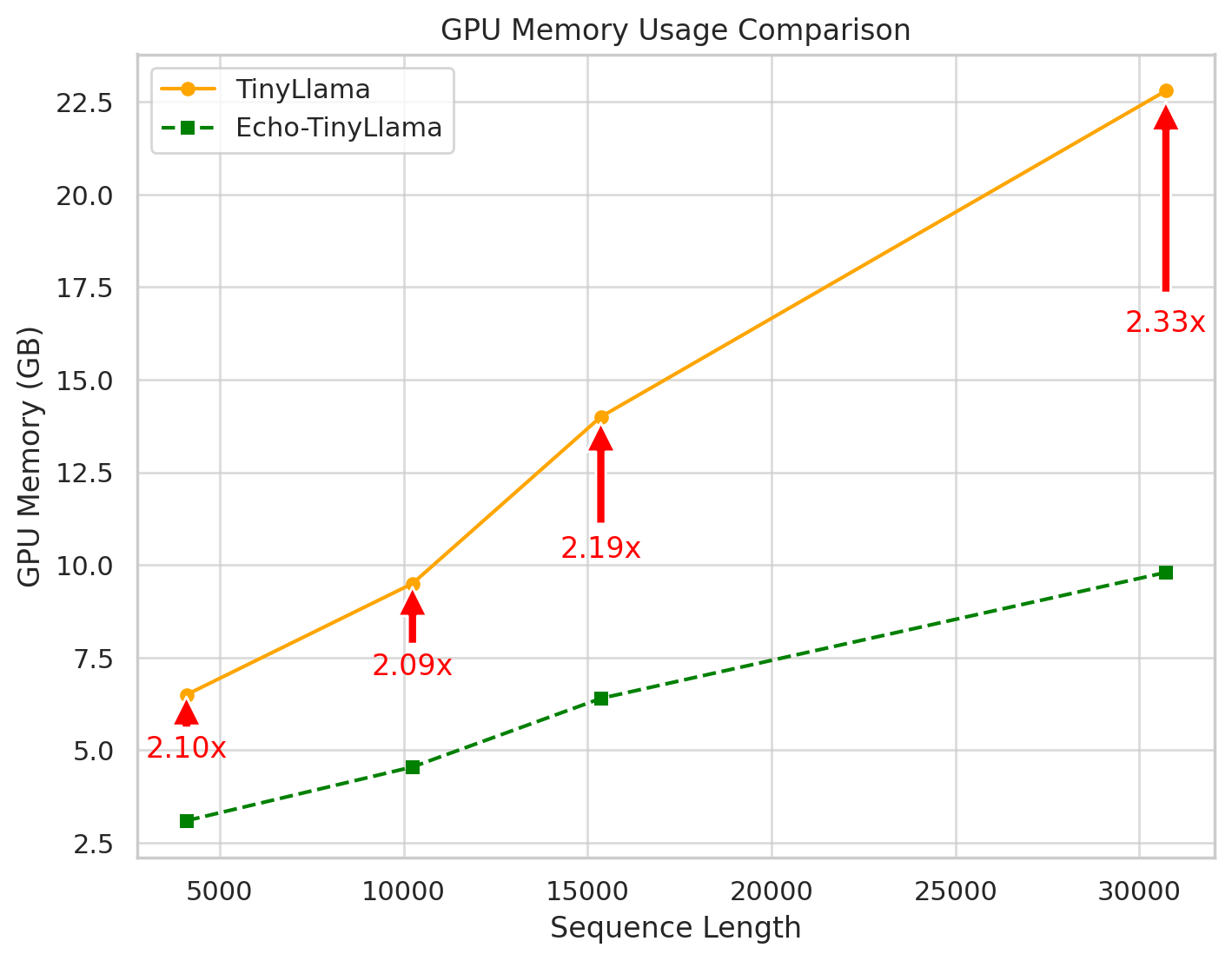}
    \caption{GPU memory usage comparison between the baseline TinyLLaMA and ECHO-TinyLLaMA across different sequence lengths. ECHO-TinyLLaMA consistently consumes nearly half the memory compared to TinyLLaMA, as indicated by the annotated reduction ratios.}
    \label{fig:gpu_memory}
\end{figure}

\noindent As shown in Figure~\ref{fig:scaling_law}, the ECHO-LLaMA models consistently achieve lower training loss than the baseline models as the model size increases. This observation demonstrates that ECHO architectures adhere to the scaling law while also offering enhanced training efficiency through their innovative design.
\noindent \textbf{GPU Memory Efficiency of Echo-TinyLLaMA-} To evaluate the efficiency of ECHO-TinyLLaMA, we compare its GPU memory consumption against the baseline TinyLLaMA across various sequence lengths. As shown in Figure~\ref{fig:gpu_memory}, ECHO-TinyLLaMA achieves a substantial reduction in memory consumption, requiring almost two times less GPU memory than TinyLLama at all tested sequence lengths. For instance, at a sequence length of 16k tokens, ECHO-TinyLLaMA consumes only 6.4 GB, whereas TinyLLaMA requires 14 GB, resulting in a \textbf{2.19x} reduction. This trend continues across longer sequences, with the reduction reaching up to \textbf{2.33x} at 32k tokens. Such improvements are particularly critical for enabling longer-context processing on resource-constrained hardware.

\begin{figure}[t]
    \centering
    \includegraphics[width=\columnwidth]{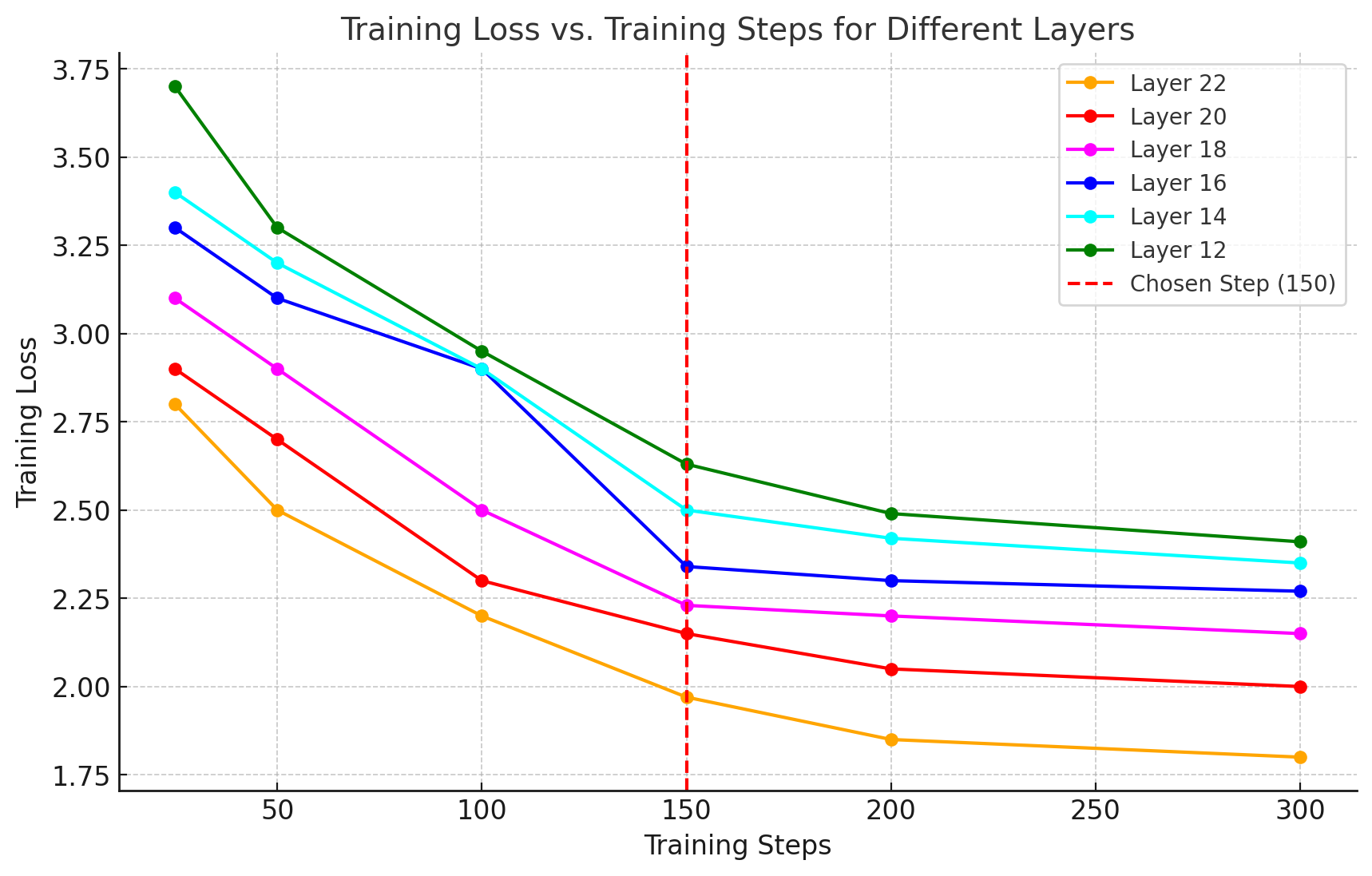}
    \caption{Training Loss vs. Training Steps for Different Layers in ECHO-TinyLLaMA. The loss decreases significantly up to 150 steps, after which the rate of improvement diminishes. This justifies our selection of $S\approx$150 training steps as the optimal point for balancing computational cost and performance.}
    \label{fig:training_loss_vs_steps}
\end{figure}

\noindent \textbf{Ablation Study: Selecting the Optimal Number of Training Steps per Stage-} we conducted an ablation study measuring the effect of different training steps $S$ on training loss. The goal was to assess at what point additional steps provide diminishing returns in loss reduction, thereby justifying our selection of 150 training steps. We evaluated training loss for ECHO-TinyLLaMA across six different layers (ranging from layer 22 to layer 12) at increasing training steps: 25, 50, 100, 150, 200, and 300. The loss values at each step were recorded to observe the rate of improvement. The plot in Figure \ref{fig:training_loss_vs_steps} illustrates the training loss progression for each layer as the number of training steps increases. We observe a sharp decrease in loss from 25 to 150 steps, indicating effective learning during this phase. However, beyond 150 steps, the slope of loss reduction significantly flattens, suggesting diminishing returns with additional training. For instance, at 150 steps, Layer 22's loss decreases from 2.8 (25 steps) → 1.97 (150 steps), a significant drop. Extending training to 200 or 300 steps results in only marginal improvements (1.85 at 200 steps, 1.80 at 300 steps), making the additional cost inefficient.

\section{Discussion}
\label{sec:discussion}
One of the main advantages of ECHO-LLaMA over YOCO models is its flexibility in choosing a balance between attention layers and cross-attention layers based on end-user needs. Unlike YOCO models, which rigidly apply shared KV caching to a fixed subset of layers, ECHO-LLaMA allows dynamic allocation of self-attention and cross-attention mechanisms. This adaptability enables fine-tuning for different deployment scenarios—whether prioritizing efficiency on resource-constrained edge devices or maximizing performance in high-compute environments. Additionally, the ability to configure attention structures per task allows ECHO-LLaMA to optimize both inference speed and long-range dependency modeling, making it a more versatile solution across diverse workloads.

Another key advantage of ECHO-LLaMA over YOCO models is its cost-effective adaptation strategy. While YOCO models require training from scratch or extensive modifications to integrate their fixed KV-sharing mechanism, ECHO-LLaMA leverages incremental adaptation to efficiently transform a pretrained LLaMA model into its structured format. This eliminates the need for expensive full-scale pretraining while maintaining—or even surpassing—baseline performance. Our experiments on TinyLLaMA demonstrate this efficiency, where a 25\% KV-sharing configuration slightly outperformed the baseline in language modeling, and a 50\% KV-sharing configuration achieved comparable performance, all while significantly improving inference speed and KV cache efficiency.

\vspace*{-.1em}

\section{Conclusion}
\label{sec:conclusion}
This work presents ECHO-LLaMA, a flexible cross layer KV sharing approach by addressing limitations of prior art such as fixed architecture and challenges in converting pretrained models to KV sharing architectures. The superiority of the proposed method is demonstrated on a range of LLaMA models ranging from 125M to 7B parameters. 
Through the incremental adaptation training strategy, we show that pretrained LLMs can be seamlessly converted into ECHO-LLaMA structures, achieving significant improvements in training efficiency with minimal performance trade-offs. The 25\% cross-decoder configuration delivers slightly improved performance, while the 50\% configuration maintains competitive performance at 7\% throughput improvement 
demonstrating the flexibility and scalability of the approach. Furthermore, pretraining from scratch highlights that ECHO-based architectures consistently achieve equal or lower training loss than baseline models, with training speeds up to 50\% faster. This makes ECHO-LLaMA architectures particularly appealing for resource-constrained environments or large-scale LLM projects requiring high computational efficiency. While the trade-off in performance for some configurations underscores the challenges of optimizing both efficiency and accuracy, our findings lay the groundwork for future innovations to efficient architectures. Future research will focus on refining the ECHO-LLaMA architecture, exploring advanced dynamic caching mechanisms, nonlinear shared KV mapping, and enhancing training strategies to fully leverage the potential of this approach.


\section{Limitations}
\label{sec:limitations}
The proposed framework relies on the availability of pre-trained models, which inherently constrains its applicability to scenarios where such models are accessible. Additionally, although the ECHO-LLaMA adaptation reduces computational overhead and improves inference speed, the reliance on shared key-value caching may introduce challenges for extremely long sequence tasks, where memory bottlenecks could still occur. Furthermore, the evaluation primarily focuses on zero/few shot performance and model efficiency, leaving broader generalization capabilities (e.g., domain adaptation or multi-task learning) unexplored.
\clearpage
\bibliography{custom}

\begin{thebibliography}{34}
\providecommand{\natexlab}[1]{#1}

\bibitem[{hug()}]{huggingfaceTinyLlama_v11}

\newblock {T}iny{L}lama/{T}iny{L}lama\_v1.1 · {H}ugging {F}ace --- huggingface.co.
\newblock \url{https://huggingface.co/TinyLlama/TinyLlama_v1.1}.
\newblock [Accessed 27-01-2025].

\bibitem[{Adnan et~al.(2024)Adnan, Arunkumar, Jain, Nair, Soloveychik, and Kamath}]{adnan2024keyformer}
Muhammad Adnan, Akhil Arunkumar, Gaurav Jain, Prashant Nair, Ilya Soloveychik, and Purushotham Kamath. 2024.
\newblock Keyformer: Kv cache reduction through key tokens selection for efficient generative inference.
\newblock \emph{Proceedings of Machine Learning and Systems}, 6:114--127.

\bibitem[{Ahmed et~al.(2023)Ahmed, Hajimolahoseini, Wen, and Liu}]{ahmed2023speeding}
Walid Ahmed, Habib Hajimolahoseini, Austin Wen, and Yang Liu. 2023.
\newblock Speeding up resnet architecture with layers targeted low rank decomposition.
\newblock \emph{arXiv preprint arXiv:2309.12412}.

\bibitem[{Austin et~al.(2021)Austin, Odena, Nye, Bosma, Michalewski, Dohan, Jiang, Cai, Terry, Le et~al.}]{austin2021program}
Jacob Austin, Augustus Odena, Maxwell Nye, Maarten Bosma, Henryk Michalewski, David Dohan, Ellen Jiang, Carrie Cai, Michael Terry, Quoc Le, et~al. 2021.
\newblock Program synthesis with large language models.
\newblock \emph{arXiv preprint arXiv:2108.07732}.

\bibitem[{Bai et~al.(2023)Bai, Bai, Chu, Cui, Dang, Deng, Fan, Ge, Han, Huang, Hui, Ji, Li, Lin, Lin, Liu, Liu, Lu, Lu, Ma, Men, Ren, Ren, Tan, Tan, Tu, Wang, Wang, Wang, Wu, Xu, Xu, Yang, Yang, Yang, Yang, Yao, Yu, Yuan, Yuan, Zhang, Zhang, Zhang, Zhang, Zhou, Zhou, Zhou, and Zhu}]{qwen}
Jinze Bai, Shuai Bai, Yunfei Chu, Zeyu Cui, Kai Dang, Xiaodong Deng, Yang Fan, Wenbin Ge, Yu~Han, Fei Huang, Binyuan Hui, Luo Ji, Mei Li, Junyang Lin, Runji Lin, Dayiheng Liu, Gao Liu, Chengqiang Lu, Keming Lu, Jianxin Ma, Rui Men, Xingzhang Ren, Xuancheng Ren, Chuanqi Tan, Sinan Tan, Jianhong Tu, Peng Wang, Shijie Wang, Wei Wang, Shengguang Wu, Benfeng Xu, Jin Xu, An~Yang, Hao Yang, Jian Yang, Shusheng Yang, Yang Yao, Bowen Yu, Hongyi Yuan, Zheng Yuan, Jianwei Zhang, Xingxuan Zhang, Yichang Zhang, Zhenru Zhang, Chang Zhou, Jingren Zhou, Xiaohuan Zhou, and Tianhang Zhu. 2023.
\newblock Qwen technical report.
\newblock \emph{arXiv preprint arXiv:2309.16609}.

\bibitem[{Beltagy et~al.(2020)Beltagy, Peters, and Cohan}]{beltagy2020longformer}
Iz~Beltagy, Matthew~E Peters, and Arman Cohan. 2020.
\newblock Longformer: The long-document transformer.
\newblock \emph{arXiv preprint arXiv:2004.05150}.

\bibitem[{Chang et~al.(2024)Chang, Wang, Wang, Wu, Yang, Zhu, Chen, Yi, Wang, Wang et~al.}]{chang2024survey}
Yupeng Chang, Xu~Wang, Jindong Wang, Yuan Wu, Linyi Yang, Kaijie Zhu, Hao Chen, Xiaoyuan Yi, Cunxiang Wang, Yidong Wang, et~al. 2024.
\newblock A survey on evaluation of large language models.
\newblock \emph{ACM Transactions on Intelligent Systems and Technology}, 15(3):1--45.

\bibitem[{Chen et~al.(2021)Chen, Tworek, Jun, Yuan, Pinto, Kaplan, Edwards, Burda, Joseph, Brockman et~al.}]{chen2021evaluating}
Mark Chen, Jerry Tworek, Heewoo Jun, Qiming Yuan, Henrique Ponde De~Oliveira Pinto, Jared Kaplan, Harri Edwards, Yuri Burda, Nicholas Joseph, Greg Brockman, et~al. 2021.
\newblock Evaluating large language models trained on code.
\newblock \emph{arXiv preprint arXiv:2107.03374}.

\bibitem[{Chowdhery et~al.(2023)Chowdhery, Narang, Devlin, Bosma, Mishra, Roberts, Barham, Chung, Sutton, Gehrmann et~al.}]{chowdhery2023palm}
Aakanksha Chowdhery, Sharan Narang, Jacob Devlin, Maarten Bosma, Gaurav Mishra, Adam Roberts, Paul Barham, Hyung~Won Chung, Charles Sutton, Sebastian Gehrmann, et~al. 2023.
\newblock Palm: Scaling language modeling with pathways.
\newblock \emph{Journal of Machine Learning Research}, 24(240):1--113.

\bibitem[{Cobbe et~al.(2021)Cobbe, Kosaraju, Bavarian, Chen, Jun, Kaiser, Plappert, Tworek, Hilton, Nakano, Hesse, and Schulman}]{cobbe2021gsm8k}
Karl Cobbe, Vineet Kosaraju, Mohammad Bavarian, Mark Chen, Heewoo Jun, Lukasz Kaiser, Matthias Plappert, Jerry Tworek, Jacob Hilton, Reiichiro Nakano, Christopher Hesse, and John Schulman. 2021.
\newblock Training verifiers to solve math word problems.
\newblock \emph{arXiv preprint arXiv:2110.14168}.

\bibitem[{Elfwing et~al.(2018)Elfwing, Uchibe, and Doya}]{elfwing2018sigmoid}
Stefan Elfwing, Eiji Uchibe, and Kenji Doya. 2018.
\newblock Sigmoid-weighted linear units for neural network function approximation in reinforcement learning.
\newblock \emph{Neural networks}, 107:3--11.

\bibitem[{Gao et~al.(2024)Gao, Tow, Abbasi, Biderman, Black, DiPofi, Foster, Golding, Hsu, Le~Noac'h, Li, McDonell, Muennighoff, Ociepa, Phang, Reynolds, Schoelkopf, Skowron, Sutawika, Tang, Thite, Wang, Wang, and Zou}]{eval-harness}
Leo Gao, Jonathan Tow, Baber Abbasi, Stella Biderman, Sid Black, Anthony DiPofi, Charles Foster, Laurence Golding, Jeffrey Hsu, Alain Le~Noac'h, Haonan Li, Kyle McDonell, Niklas Muennighoff, Chris Ociepa, Jason Phang, Laria Reynolds, Hailey Schoelkopf, Aviya Skowron, Lintang Sutawika, Eric Tang, Anish Thite, Ben Wang, Kevin Wang, and Andy Zou. 2024.
\newblock \href {https://doi.org/10.5281/zenodo.12608602} {A framework for few-shot language model evaluation}.

\bibitem[{Hajimolahoseini et~al.(2023)Hajimolahoseini, Ahmed, and Liu}]{hajimolahoseini2023training}
Habib Hajimolahoseini, Walid Ahmed, and Yang Liu. 2023.
\newblock Training acceleration of low-rank decomposed networks using sequential freezing and rank quantization.
\newblock \emph{arXiv preprint arXiv:2309.03824}.

\bibitem[{Hendrycks et~al.(2021{\natexlab{a}})Hendrycks, Burns, Basart, Critch, Li, Song, and Steinhardt}]{hendrycks2021ethics}
Dan Hendrycks, Collin Burns, Steven Basart, Andrew Critch, Jerry Li, Dawn Song, and Jacob Steinhardt. 2021{\natexlab{a}}.
\newblock Aligning ai with shared human values.
\newblock \emph{Proceedings of the International Conference on Learning Representations (ICLR)}.

\bibitem[{Hendrycks et~al.(2021{\natexlab{b}})Hendrycks, Burns, Basart, Zou, Mazeika, Song, and Steinhardt}]{hendryckstest2021}
Dan Hendrycks, Collin Burns, Steven Basart, Andy Zou, Mantas Mazeika, Dawn Song, and Jacob Steinhardt. 2021{\natexlab{b}}.
\newblock Measuring massive multitask language understanding.
\newblock \emph{Proceedings of the International Conference on Learning Representations (ICLR)}.

\bibitem[{Hendrycks et~al.(2021{\natexlab{c}})Hendrycks, Burns, Kadavath, Arora, Basart, Tang, Song, and Steinhardt}]{hendrycks2021measuring}
Dan Hendrycks, Collin Burns, Saurav Kadavath, Akul Arora, Steven Basart, Eric Tang, Dawn Song, and Jacob Steinhardt. 2021{\natexlab{c}}.
\newblock Measuring mathematical problem solving with the math dataset.
\newblock \emph{arXiv preprint arXiv:2103.03874}.

\bibitem[{Hoffmann et~al.(2022)Hoffmann, Borgeaud, Mensch, Buchatskaya, Cai, Rutherford, Casas, Hendricks, Welbl, Clark et~al.}]{hoffmann2022training}
Jordan Hoffmann, Sebastian Borgeaud, Arthur Mensch, Elena Buchatskaya, Trevor Cai, Eliza Rutherford, Diego de~Las Casas, Lisa~Anne Hendricks, Johannes Welbl, Aidan Clark, et~al. 2022.
\newblock Training compute-optimal large language models.
\newblock \emph{arXiv preprint arXiv:2203.15556}.

\bibitem[{Huang et~al.(2023)Huang, Bai, Zhu, Zhang, Zhang, Su, Liu, Lv, Zhang, Lei, Fu, Sun, and He}]{huang2023ceval}
Yuzhen Huang, Yuzhuo Bai, Zhihao Zhu, Junlei Zhang, Jinghan Zhang, Tangjun Su, Junteng Liu, Chuancheng Lv, Yikai Zhang, Jiayi Lei, Yao Fu, Maosong Sun, and Junxian He. 2023.
\newblock C-eval: A multi-level multi-discipline chinese evaluation suite for foundation models.
\newblock \emph{arXiv preprint arXiv:2305.08322}.

\bibitem[{Javadi et~al.(2023)Javadi, Ahmed, Hajimolahoseini, Ataiefard, Hassanpour, Asani, Wen, Awad, Liu, and Liu}]{javadi2023gqkva}
Farnoosh Javadi, Walid Ahmed, Habib Hajimolahoseini, Foozhan Ataiefard, Mohammad Hassanpour, Saina Asani, Austin Wen, Omar~Mohamed Awad, Kangling Liu, and Yang Liu. 2023.
\newblock Gqkva: Efficient pre-training of transformers by grouping queries, keys, and values.
\newblock \emph{arXiv preprint arXiv:2311.03426}.

\bibitem[{Kirkpatrick et~al.(2017)Kirkpatrick, Pascanu, Rabinowitz, Veness, Desjardins, Rusu, Milan, Quan, Ramalho, Grabska-Barwinska et~al.}]{kirkpatrick2017overcoming}
James Kirkpatrick, Razvan Pascanu, Neil Rabinowitz, Joel Veness, Guillaume Desjardins, Andrei~A Rusu, Kieran Milan, John Quan, Tiago Ramalho, Agnieszka Grabska-Barwinska, et~al. 2017.
\newblock Overcoming catastrophic forgetting in neural networks.
\newblock \emph{Proceedings of the national academy of sciences}, 114(13):3521--3526.

\bibitem[{Lee et~al.(2024)Lee, Lee, Seo, and Sim}]{lee2024infinigen}
Wonbeom Lee, Jungi Lee, Junghwan Seo, and Jaewoong Sim. 2024.
\newblock $\{$InfiniGen$\}$: Efficient generative inference of large language models with dynamic $\{$KV$\}$ cache management.
\newblock In \emph{18th USENIX Symposium on Operating Systems Design and Implementation (OSDI 24)}, pages 155--172.

\bibitem[{Liao and Vargas(2024)}]{liao2024beyond}
Bingli Liao and Danilo~Vasconcellos Vargas. 2024.
\newblock Beyond kv caching: Shared attention for efficient llms.
\newblock \emph{arXiv preprint arXiv:2407.12866}.

\bibitem[{Makridakis et~al.(2023)Makridakis, Petropoulos, and Kang}]{makridakis2023large}
Spyros Makridakis, Fotios Petropoulos, and Yanfei Kang. 2023.
\newblock Large language models: Their success and impact.
\newblock \emph{Forecasting}, 5(3):536--549.

\bibitem[{Milano et~al.(2023)Milano, McGrane, and Leonelli}]{milano2023large}
Silvia Milano, Joshua~A McGrane, and Sabina Leonelli. 2023.
\newblock Large language models challenge the future of higher education.
\newblock \emph{Nature Machine Intelligence}, 5(4):333--334.

\bibitem[{Pope et~al.(2023)Pope, Douglas, Chowdhery, Devlin, Bradbury, Heek, Xiao, Agrawal, and Dean}]{pope2023efficiently}
Reiner Pope, Sholto Douglas, Aakanksha Chowdhery, Jacob Devlin, James Bradbury, Jonathan Heek, Kefan Xiao, Shivani Agrawal, and Jeff Dean. 2023.
\newblock Efficiently scaling transformer inference.
\newblock \emph{Proceedings of Machine Learning and Systems}, 5:606--624.

\bibitem[{Rajabzadeh et~al.(2024)Rajabzadeh, Jafari, Sharma, Jami, Kwon, Ghodsi, Chen, and Rezagholizadeh}]{rajabzadeh2024echoatt}
Hossein Rajabzadeh, Aref Jafari, Aman Sharma, Benyamin Jami, Hyock~Ju Kwon, Ali Ghodsi, Boxing Chen, and Mehdi Rezagholizadeh. 2024.
\newblock Echoatt: Attend, copy, then adjust for more efficient large language models.
\newblock \emph{arXiv preprint arXiv:2409.14595}.

\bibitem[{Sun et~al.(2024)Sun, Dong, Zhu, Huang, Wang, Ma, Zhang, Wang, and Wei}]{sun2024you}
Yutao Sun, Li~Dong, Yi~Zhu, Shaohan Huang, Wenhui Wang, Shuming Ma, Quanlu Zhang, Jianyong Wang, and Furu Wei. 2024.
\newblock You only cache once: Decoder-decoder architectures for language models.
\newblock \emph{arXiv preprint arXiv:2405.05254}.

\bibitem[{Tang et~al.(2024)Tang, Lin, Lin, Han, Hong, Yao, and Wang}]{tang2024razorattention}
Hanlin Tang, Yang Lin, Jing Lin, Qingsen Han, Shikuan Hong, Yiwu Yao, and Gongyi Wang. 2024.
\newblock Razorattention: Efficient kv cache compression through retrieval heads.
\newblock \emph{arXiv preprint arXiv:2407.15891}.

\bibitem[{Touvron et~al.(2023)Touvron, Martin, Stone, Albert, Almahairi, Babaei, Bashlykov, Batra, Bhargava, Bhosale et~al.}]{touvron2023llama}
Hugo Touvron, Louis Martin, Kevin Stone, Peter Albert, Amjad Almahairi, Yasmine Babaei, Nikolay Bashlykov, Soumya Batra, Prajjwal Bhargava, Shruti Bhosale, et~al. 2023.
\newblock Llama 2: Open foundation and fine-tuned chat models.
\newblock \emph{arXiv preprint arXiv:2307.09288}.

\bibitem[{Yang et~al.(2023)Yang, Wang, Shen, Panda, and Kim}]{yang2023gated}
Songlin Yang, Bailin Wang, Yikang Shen, Rameswar Panda, and Yoon Kim. 2023.
\newblock Gated linear attention transformers with hardware-efficient training.
\newblock \emph{arXiv preprint arXiv:2312.06635}.

\bibitem[{Zhang and Sennrich(2019)}]{zhang2019root}
Biao Zhang and Rico Sennrich. 2019.
\newblock Root mean square layer normalization.
\newblock \emph{Advances in Neural Information Processing Systems}, 32.

\bibitem[{Zhang et~al.(2024)Zhang, Ladhak, Durmus, Liang, McKeown, and Hashimoto}]{zhang2024benchmarking}
Tianyi Zhang, Faisal Ladhak, Esin Durmus, Percy Liang, Kathleen McKeown, and Tatsunori~B Hashimoto. 2024.
\newblock Benchmarking large language models for news summarization.
\newblock \emph{Transactions of the Association for Computational Linguistics}, 12:39--57.

\bibitem[{Zhang et~al.(2023)Zhang, Sheng, Zhou, Chen, Zheng, Cai, Song, Tian, R{\'e}, Barrett et~al.}]{zhang2023h2o}
Zhenyu Zhang, Ying Sheng, Tianyi Zhou, Tianlong Chen, Lianmin Zheng, Ruisi Cai, Zhao Song, Yuandong Tian, Christopher R{\'e}, Clark Barrett, et~al. 2023.
\newblock H2o: Heavy-hitter oracle for efficient generative inference of large language models.
\newblock \emph{Advances in Neural Information Processing Systems}, 36:34661--34710.

\bibitem[{Zheng et~al.(2024)Zheng, Zhang, Zhang, Ye, Luo, Feng, and Ma}]{zheng2024llamafactory}
Yaowei Zheng, Richong Zhang, Junhao Zhang, Yanhan Ye, Zheyan Luo, Zhangchi Feng, and Yongqiang Ma. 2024.
\newblock \href {http://arxiv.org/abs/2403.13372} {Llamafactory: Unified efficient fine-tuning of 100+ language models}.
\newblock In \emph{Proceedings of the 62nd Annual Meeting of the Association for Computational Linguistics (Volume 3: System Demonstrations)}, Bangkok, Thailand. Association for Computational Linguistics.

\end{thebibliography}
\clearpage
\appendix Appedix

\section{Training Efficiency and Performance Comparison of ECHO-LLaMA Models}
\begin{figure*}[t]
    \centering
    \includegraphics[width=0.9\textwidth]{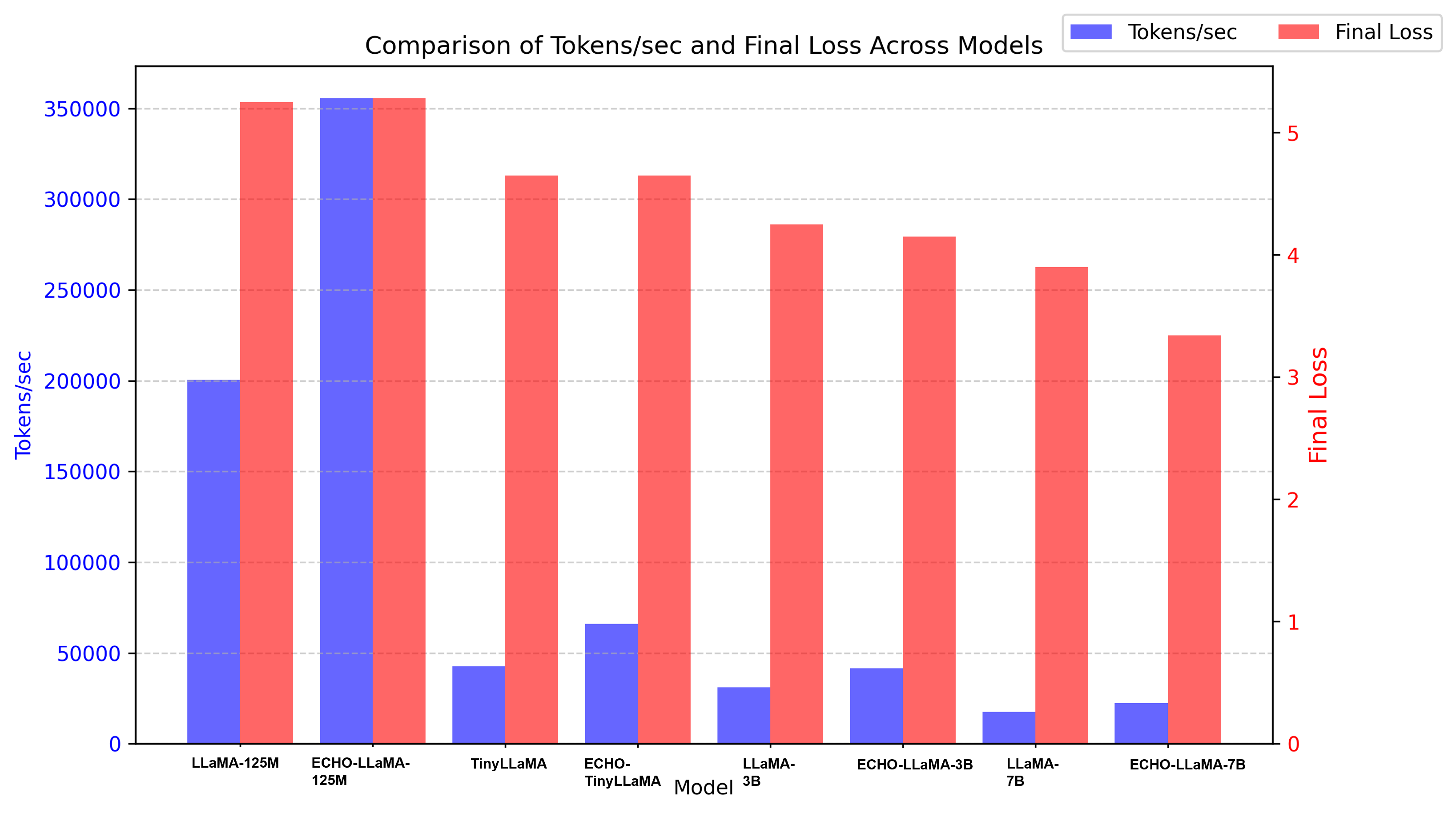}
    \caption{Comparison of training throughput (Tokens/sec) and final loss for ECHO and non-ECHO models on NPU-910B. ECHO-LLaMA versions consistently achieve lower loss with higher or comparable Tokens/sec speed.}
    \label{fig:npu_tokens_loss}
\end{figure*}

 To further illustrate the efficiency of the ECHO-LLaMA models, we compare the training throughput (Tokens/sec) and the final loss for different model sizes. Figure~\ref{fig:npu_tokens_loss} shows the performance comparison across all four LLM sizes (125M, 1.1B, 3B, and 7B) and their ECHO-LLaMA counterparts. The results demonstrate that the ECHO-LLaMA models consistently achieve lower final loss compared to the non-ECHO versions while achieving higher or comparable throughput (Tokens/sec). For instance, ECHO-LLaMA-125M improves the training throughput by a significant margin while maintaining a comparable loss, and ECHO-LLaMA-7B reduces the loss substantially with a moderate improvement in throughput. These findings highlight the effectiveness of the ECHO structure in achieving better optimization efficiency during training.

\section{Performance Evaluation of ECHO-LLaMA Structure During Pretraining from Scratch}
In this section, we evaluate the performance of our ECHO-LLaMA structure during pretraining from scratch for different sizes of large language models (LLMs): 125M, 1.1B, 3B, and 7B. The experiments are conducted on two hardware platforms: GPU (V100) and NPU (Ascend-910B). For each LLM size, we analyze the training dynamics using three types of plots:
\begin{enumerate}
    \item \textbf{Train Loss per Steps}: Tracks the training loss as a function of the number of training steps.
    \item \textbf{Train Loss per Train Time}: Evaluates the relationship between training loss and the elapsed training time.
    \item \textbf{Train Loss per Number of Train Tokens}: Measures how efficiently the model learns as a function of the number of processed tokens.
\end{enumerate}

\begin{figure*}[ht]
    \centering
    \subsection*{LLaMA-125M Model Training Results}
    \begin{tabular}{ccc}
        \includegraphics[width=0.3\textwidth]{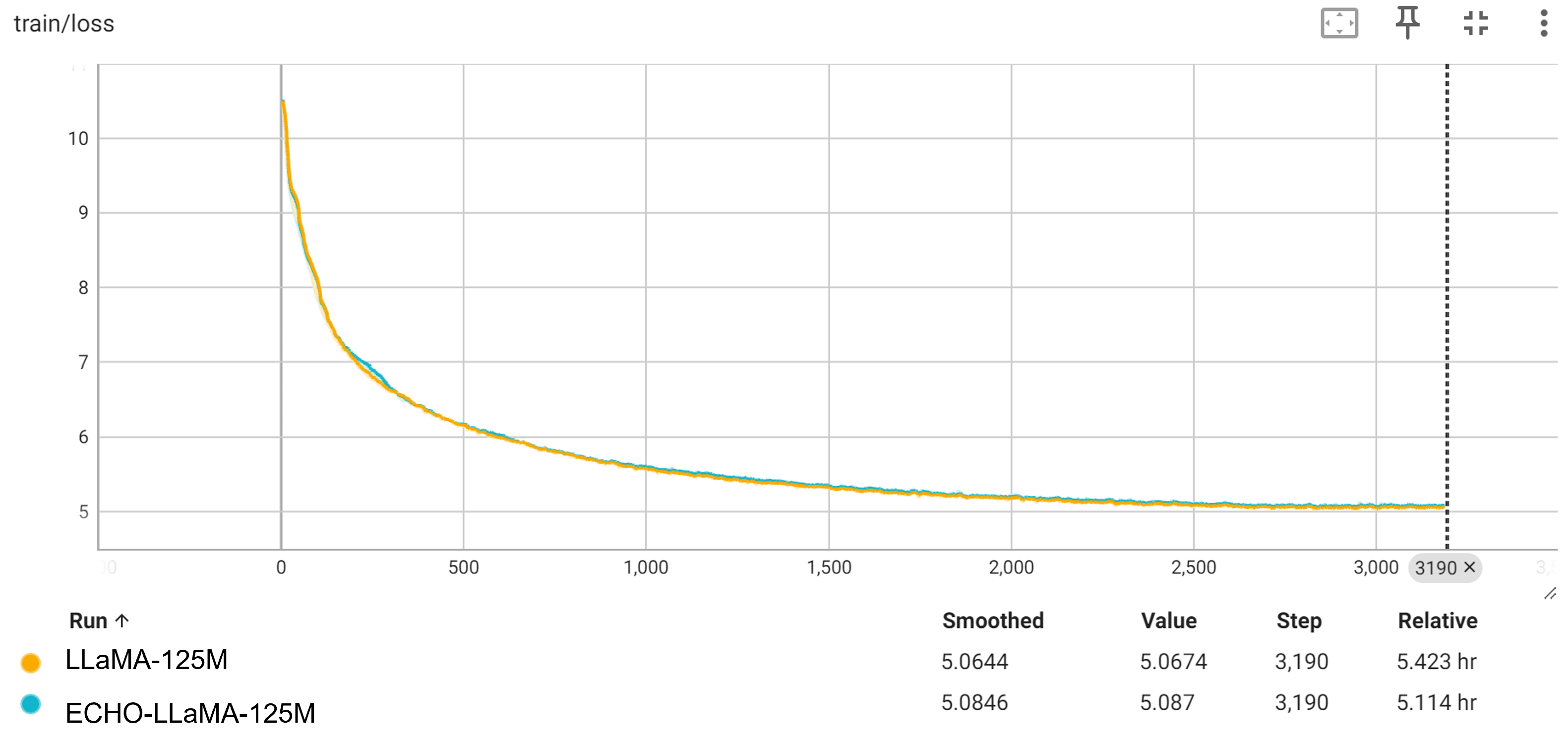} &
        \includegraphics[width=0.3\textwidth]{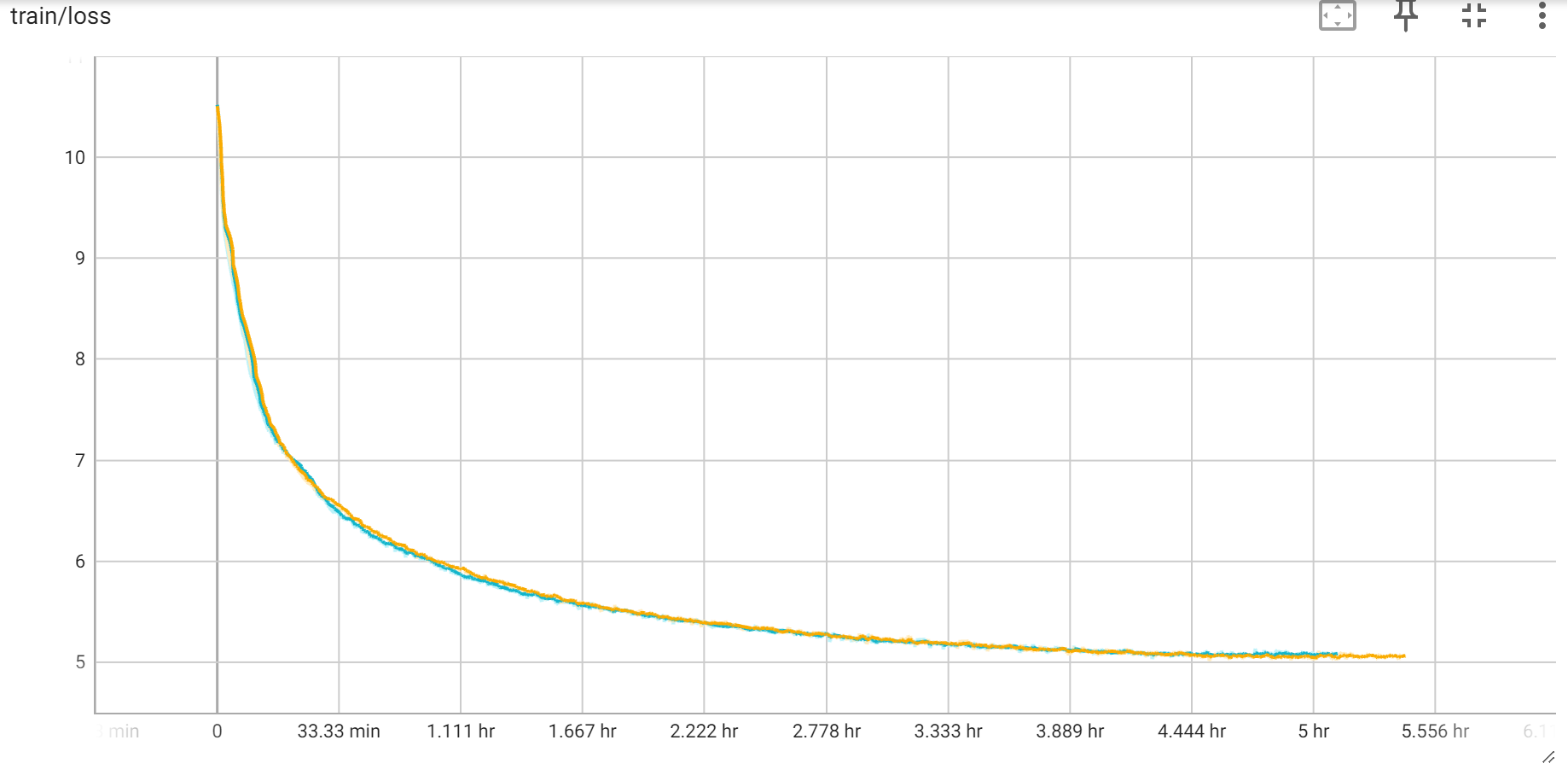} &
        \includegraphics[width=0.3\textwidth]{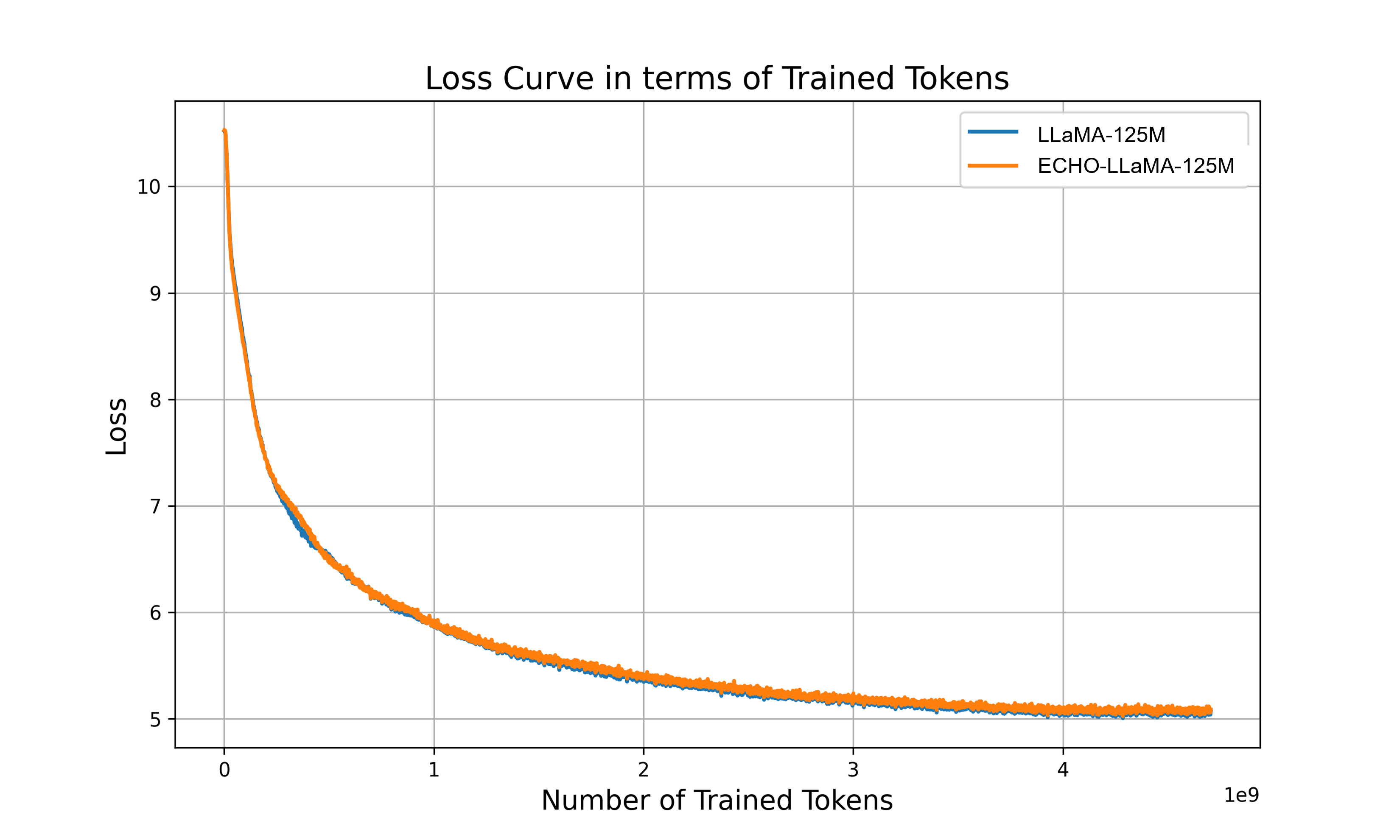} \\
        \multicolumn{3}{c}{\textbf{GPU (V100) Results}} \\
        \includegraphics[width=0.3\textwidth]{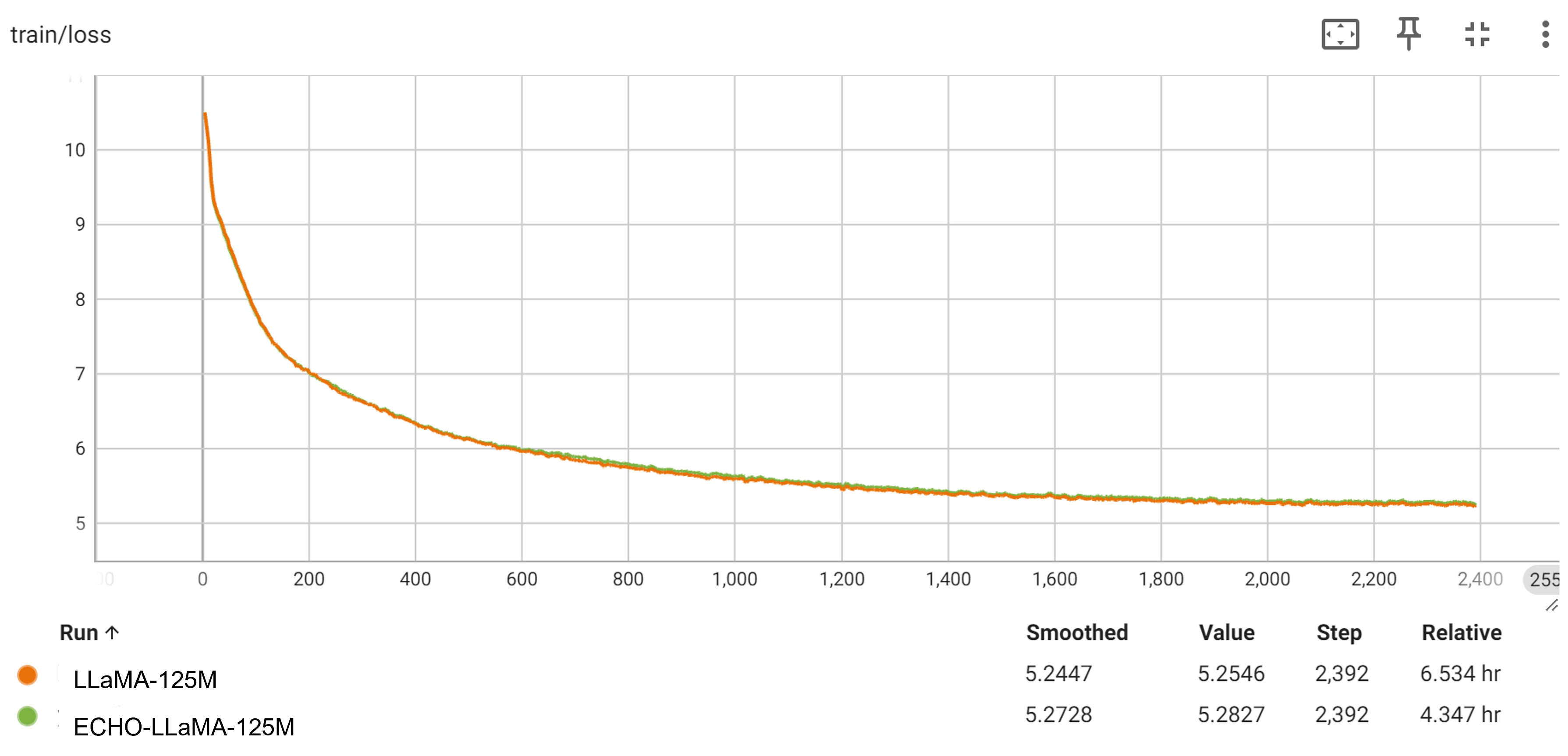} &
        \includegraphics[width=0.3\textwidth]{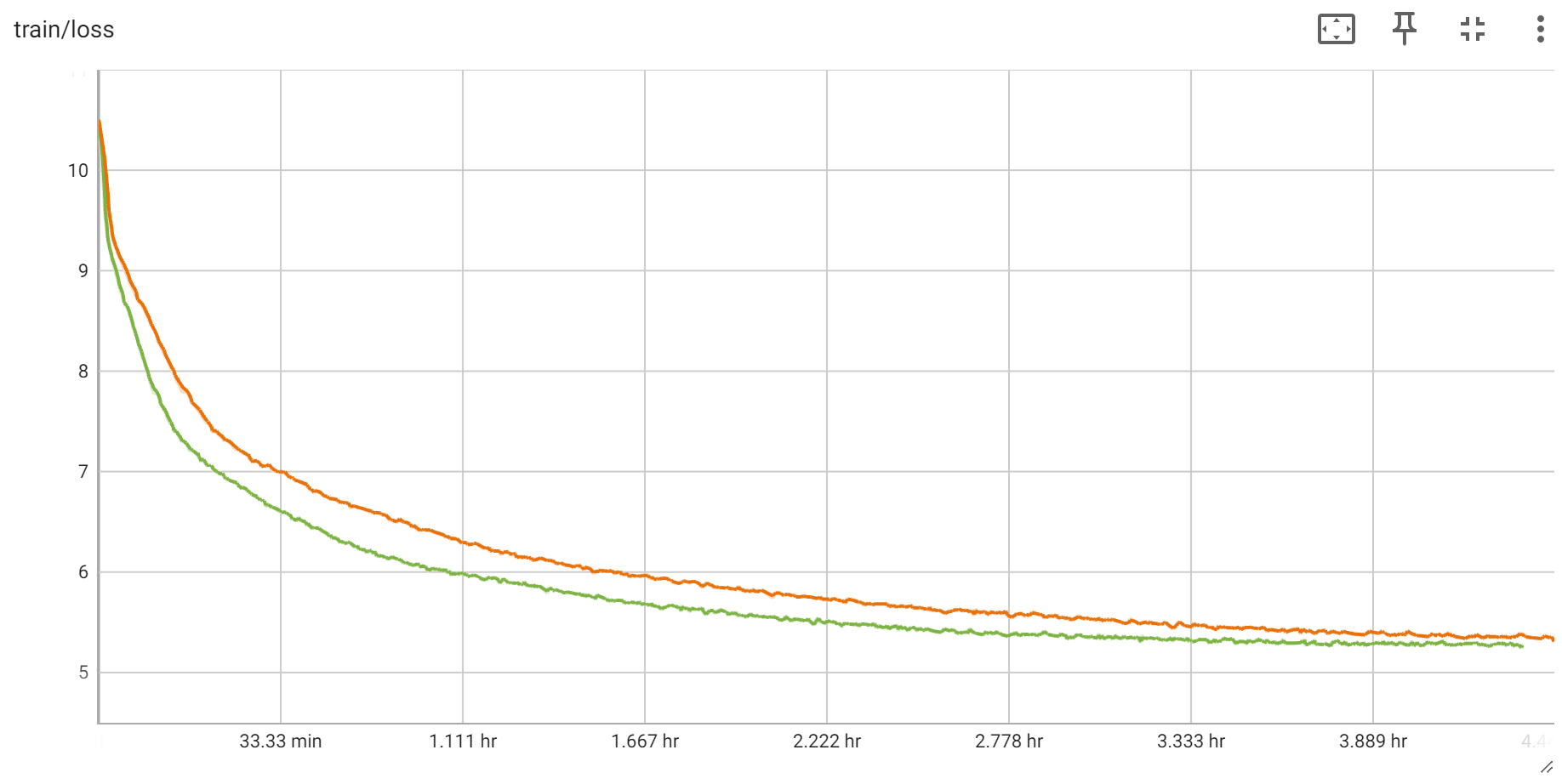} &
        \includegraphics[width=0.3\textwidth]{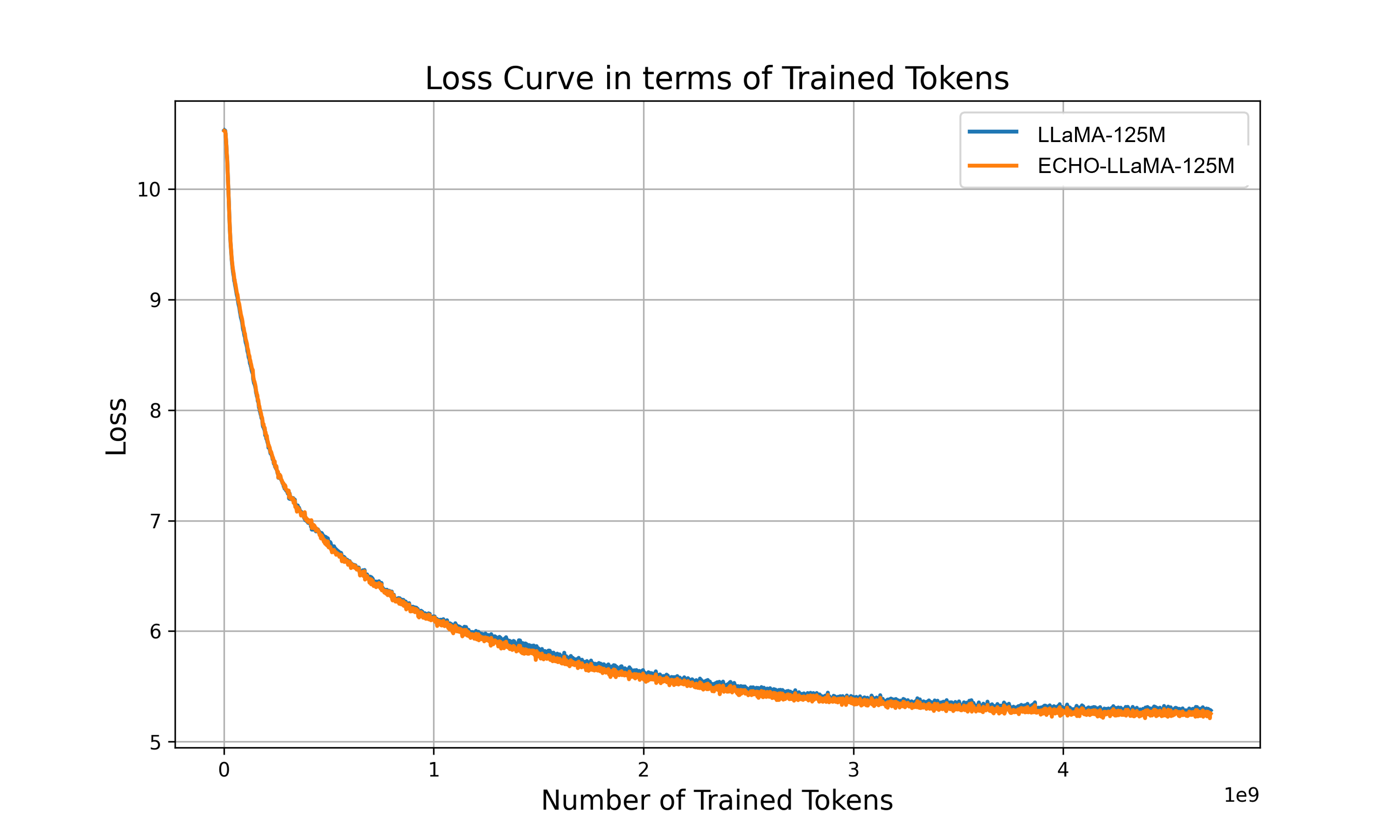} \\
        \multicolumn{3}{c}{\textbf{NPU (910B) Results}}
    \end{tabular}
    \caption{Training results for the LLaMA-125M LLM on GPU (V100) and NPU (910B). The columns represent (1) Train Loss vs. Steps, (2) Train Loss vs. Train Time, and (3) Train Loss vs. Number of Train Tokens.}
    \label{fig:125M_results}
\end{figure*}

\begin{figure*}[t]
    \centering
    \subsection*{TinyLLaMA-1.1B Model Training Results}
    \begin{tabular}{ccc}
        \includegraphics[width=0.3\textwidth]{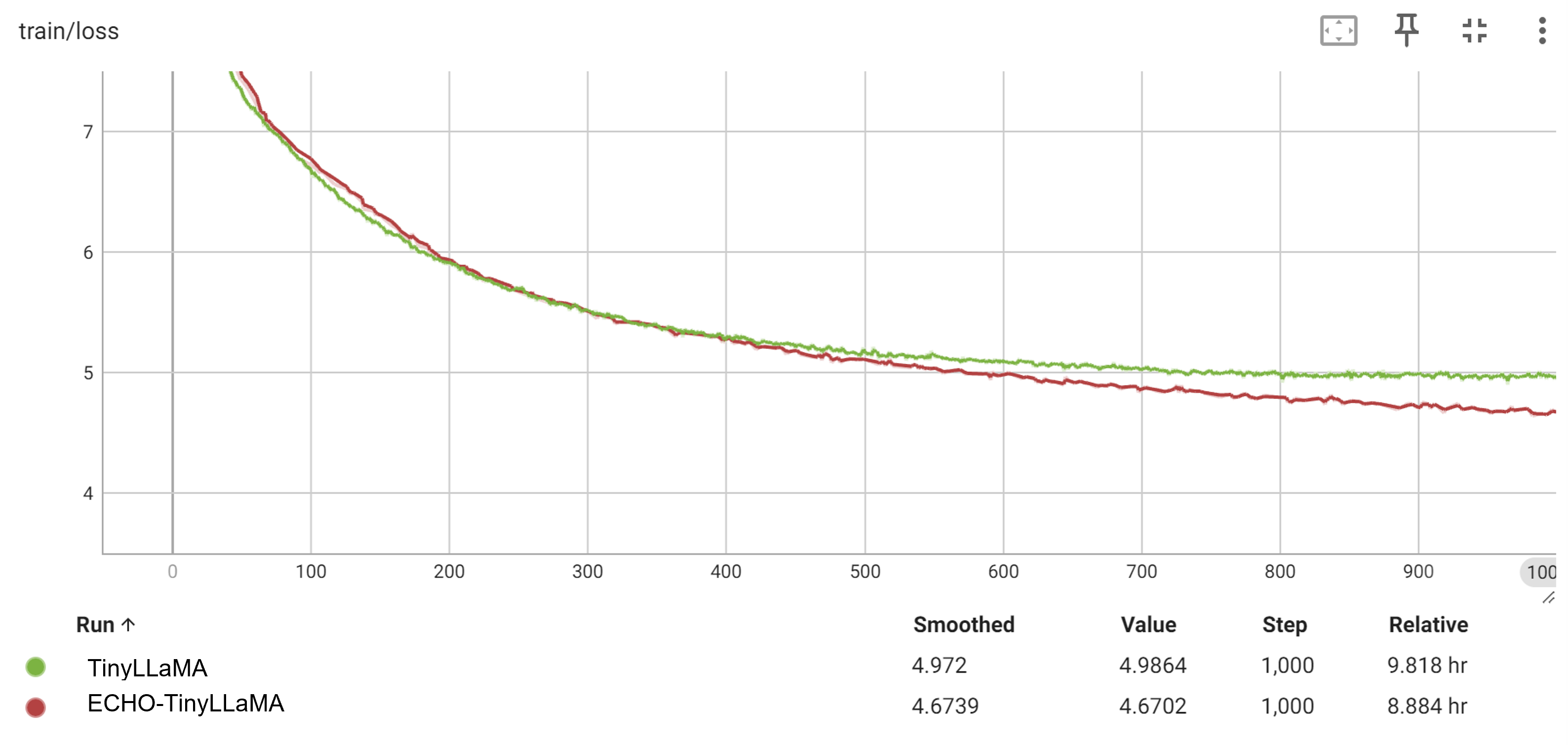} &
        \includegraphics[width=0.3\textwidth]{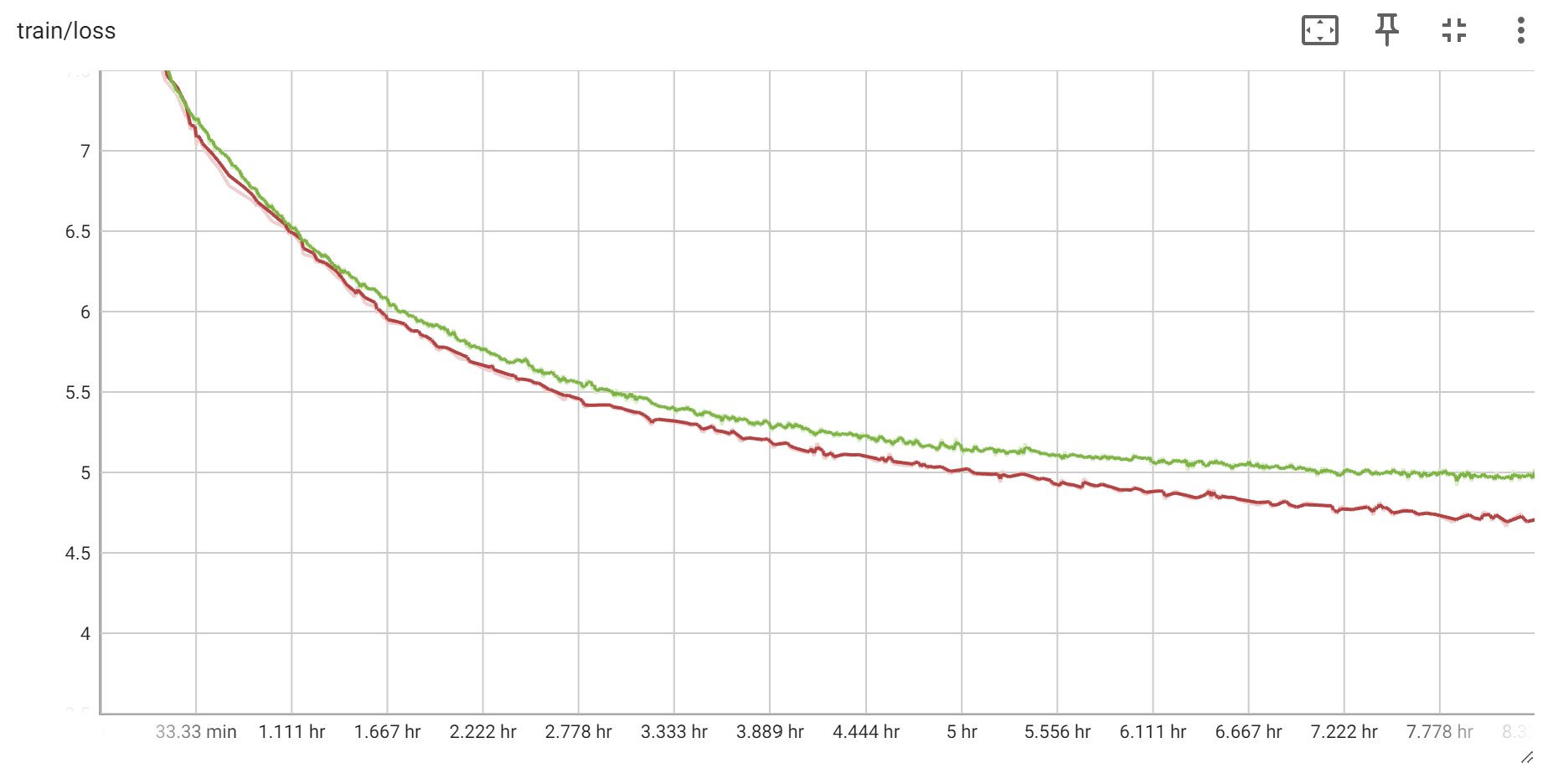} &
        \includegraphics[width=0.3\textwidth]{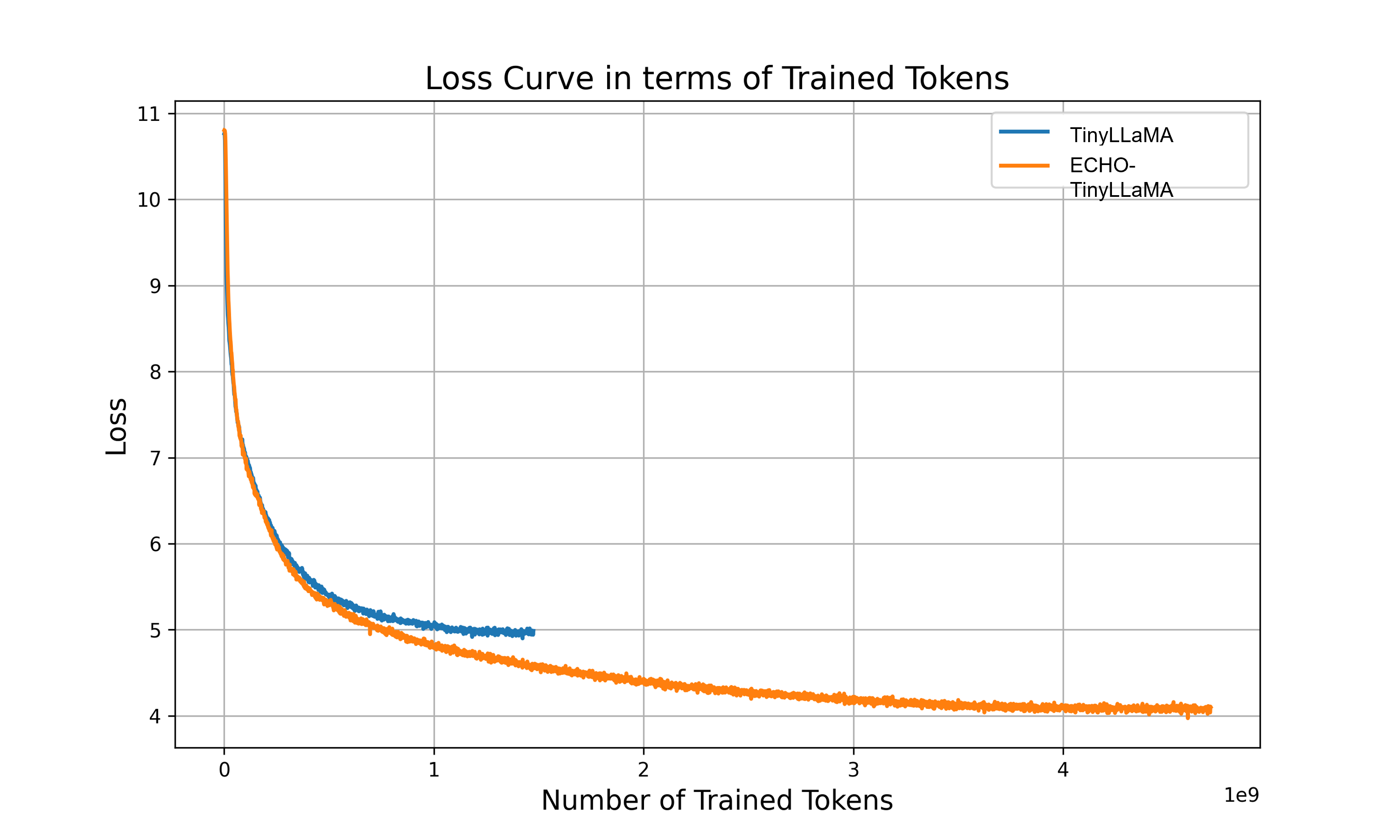} \\
        \multicolumn{3}{c}{\textbf{GPU (V100) Results}} \\
        \includegraphics[width=0.3\textwidth]{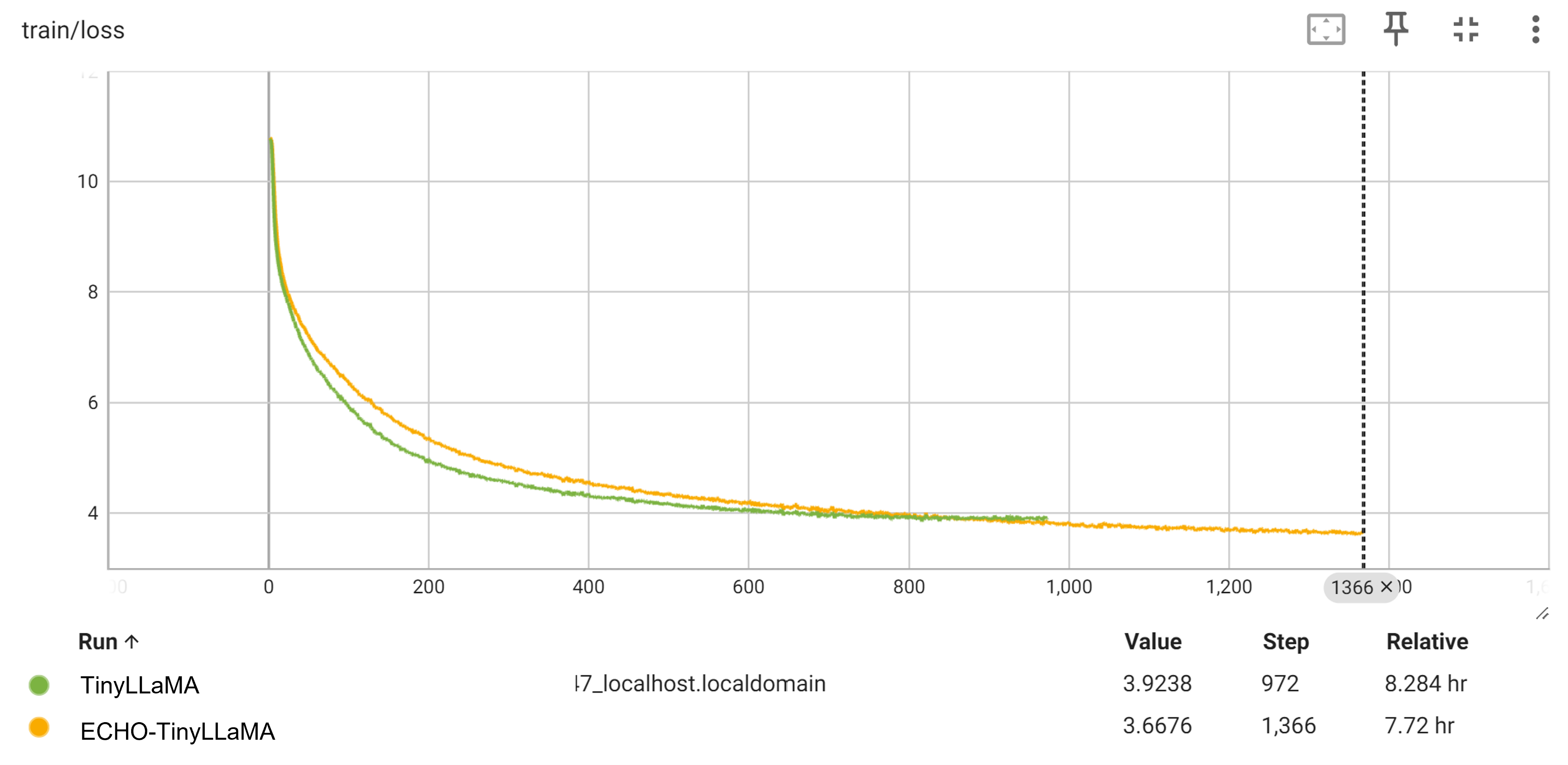} &
        \includegraphics[width=0.3\textwidth]{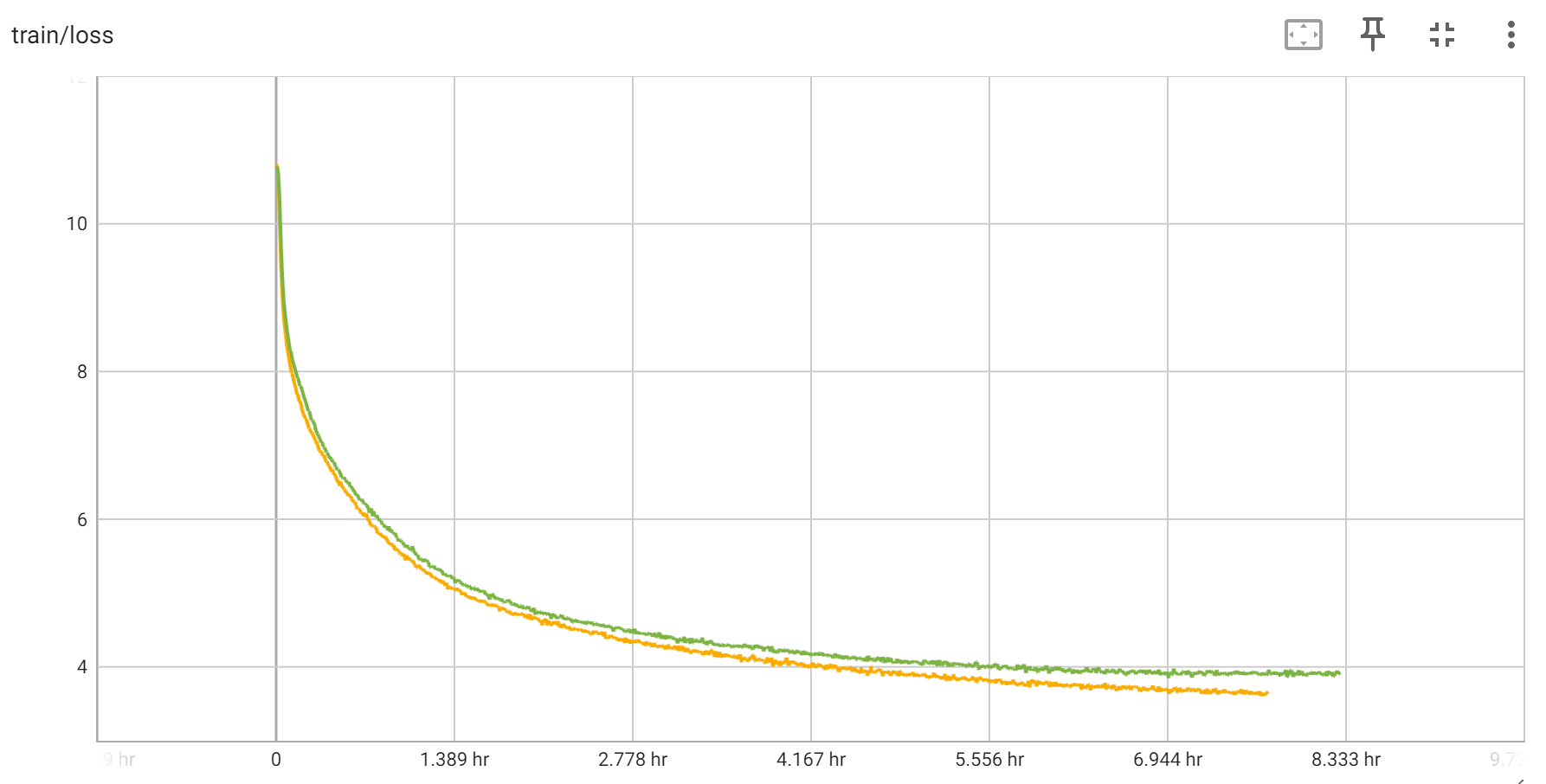} &
        \includegraphics[width=0.3\textwidth]{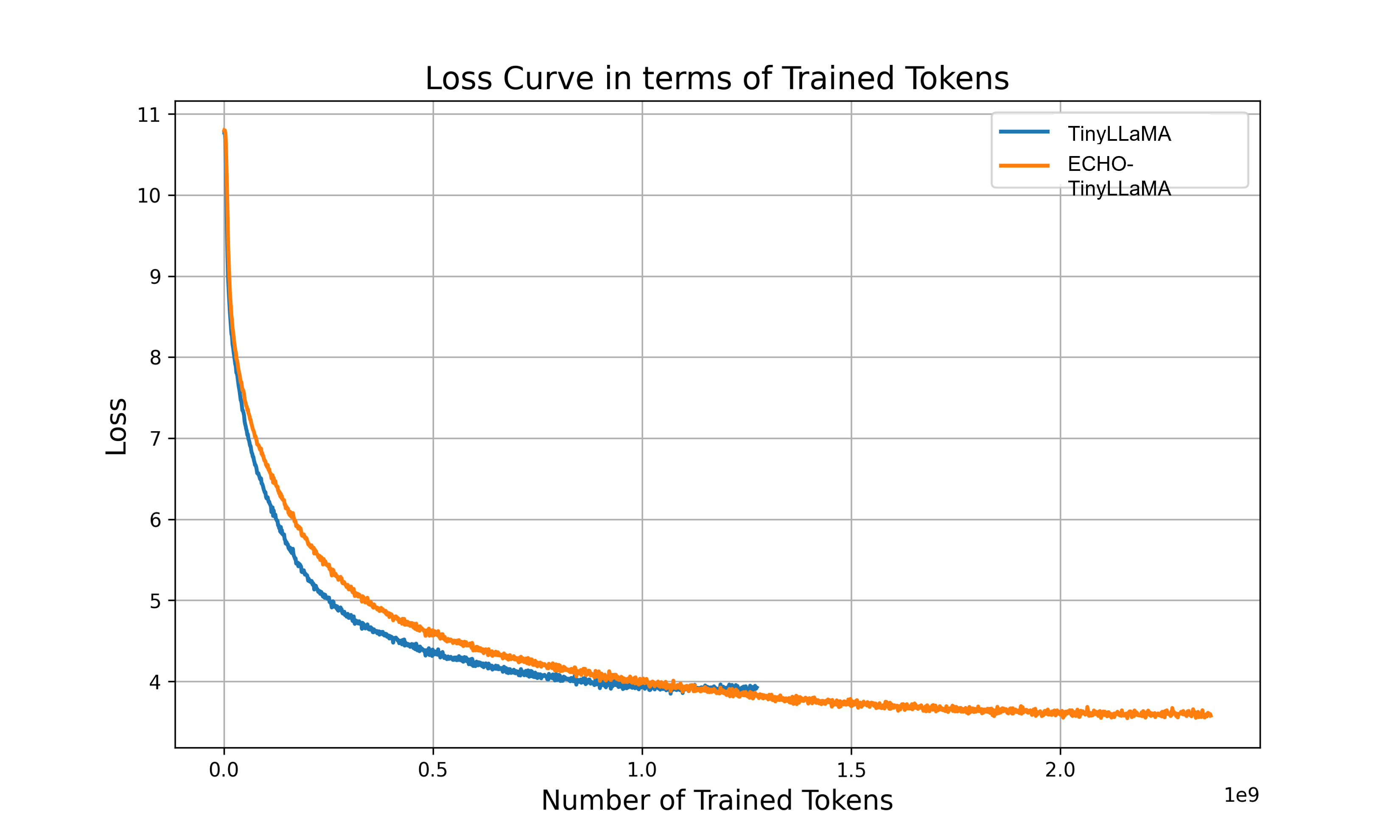} \\
        \multicolumn{3}{c}{\textbf{NPU (910B) Results}}
    \end{tabular}
    \caption{Training results for TinyLLaMA-1.1B LLM on GPU (V100) and NPU (910B). The columns represent (1) Train Loss vs. Steps, (2) Train Loss vs. Train Time, and (3) Train Loss vs. Number of Train Tokens.}
    \label{fig:1.1B_results}
\end{figure*}

\begin{figure*}[ht]
    \centering
    \subsection*{LLaMA-3B Model Training Results}
    \begin{tabular}{ccc}
        \includegraphics[width=0.3\textwidth]{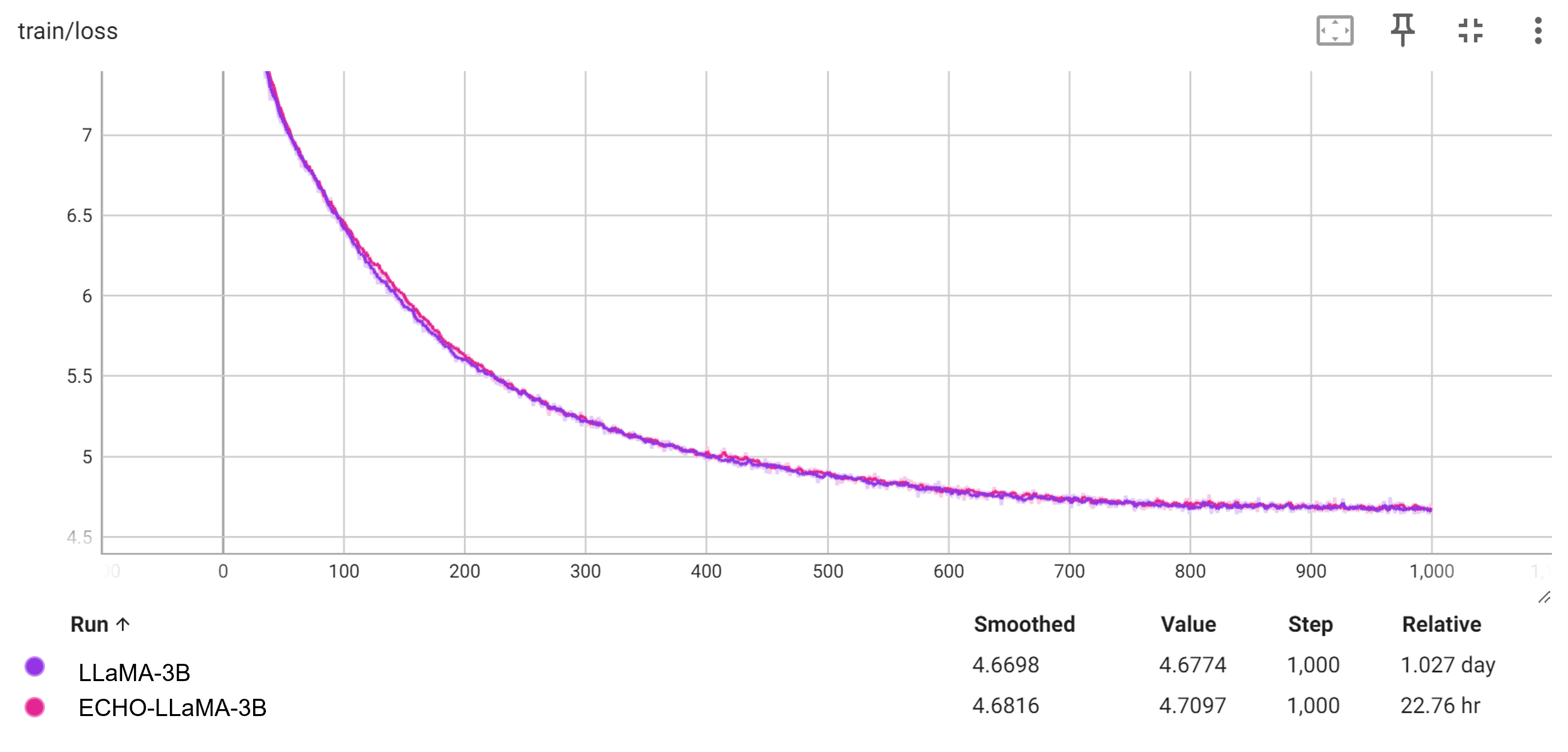} &
        \includegraphics[width=0.3\textwidth]{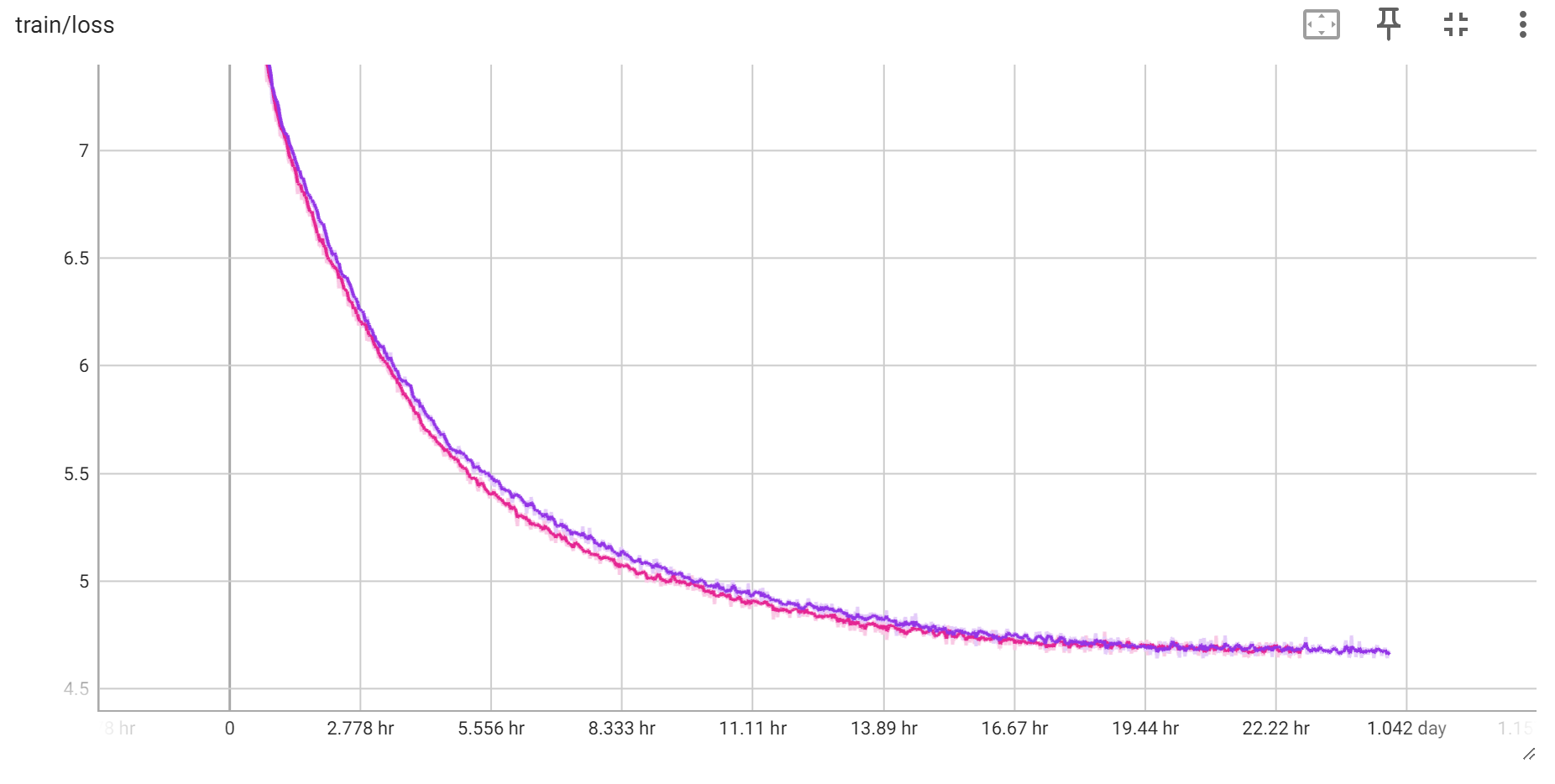} &
        \includegraphics[width=0.3\textwidth]{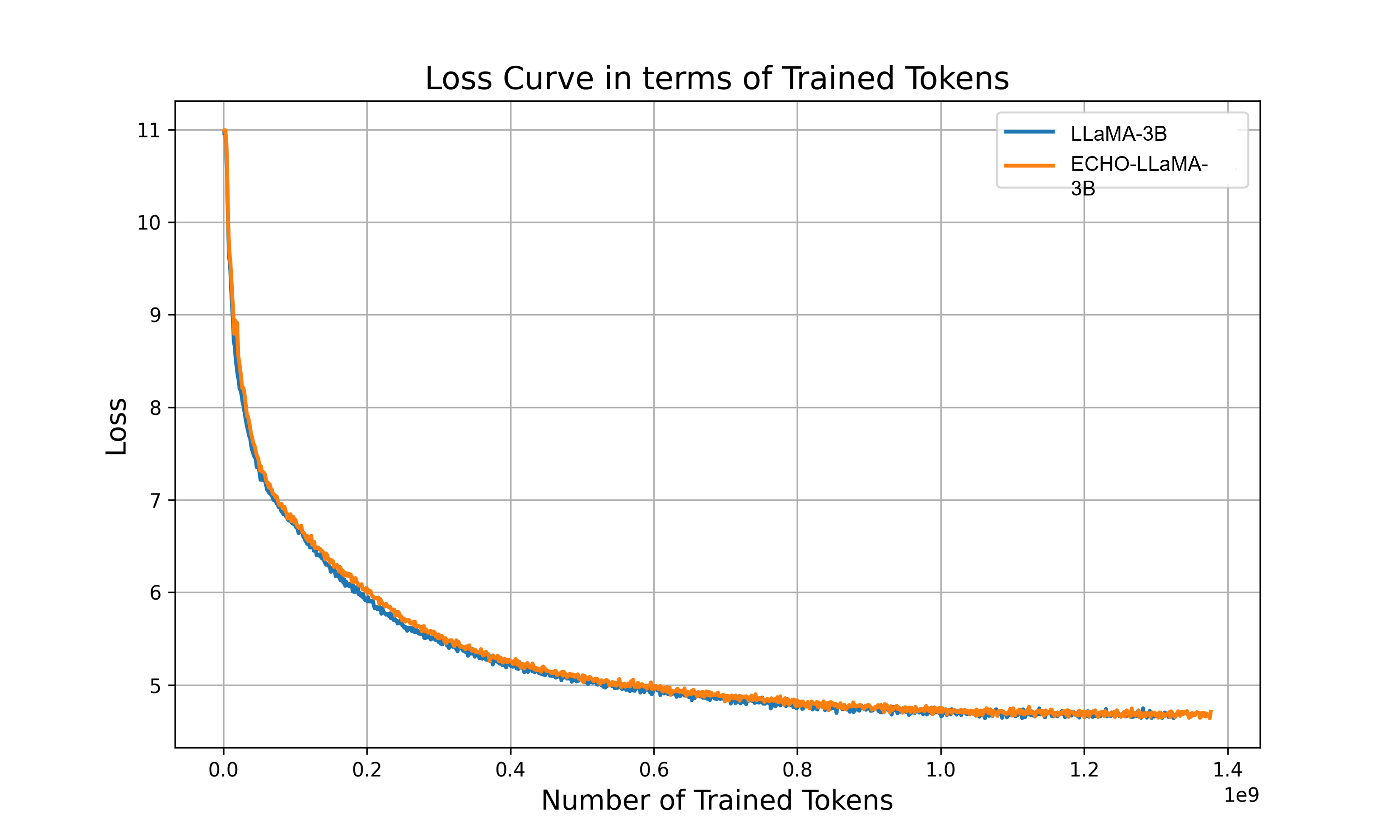} \\
        \multicolumn{3}{c}{\textbf{GPU (V100) Results}} \\
        \includegraphics[width=0.3\textwidth]{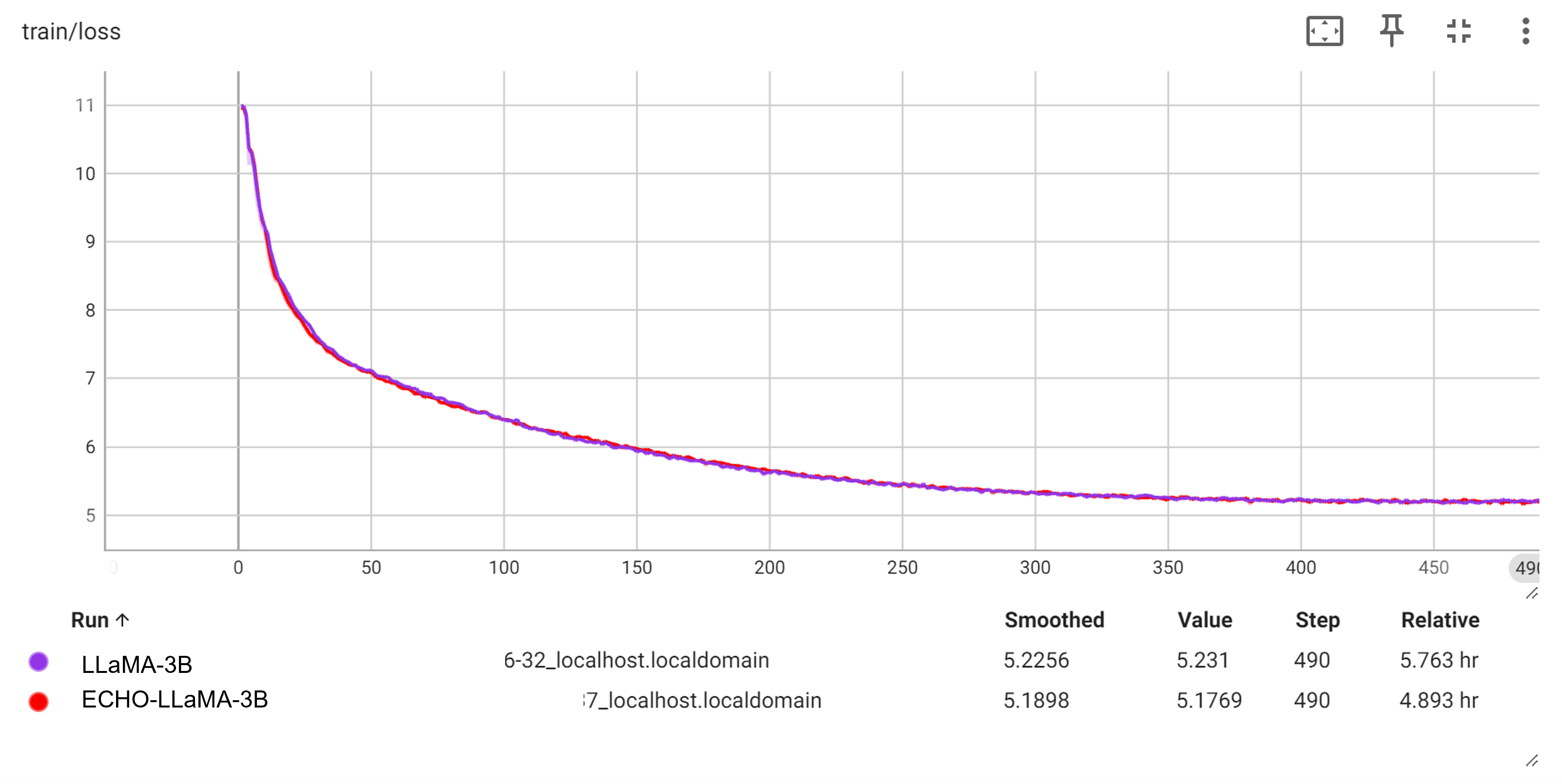} &
        \includegraphics[width=0.3\textwidth]{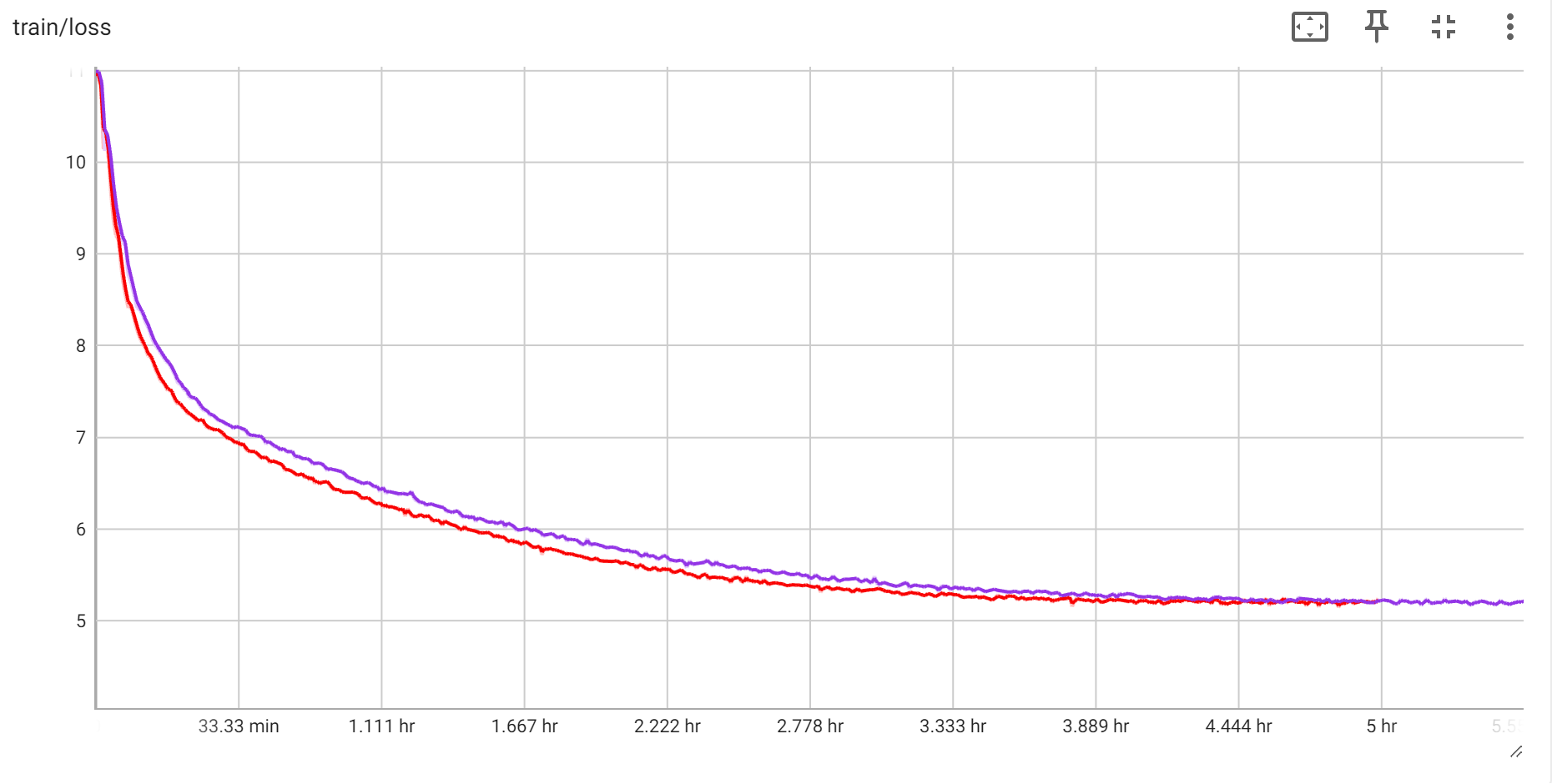} &
        \includegraphics[width=0.3\textwidth]{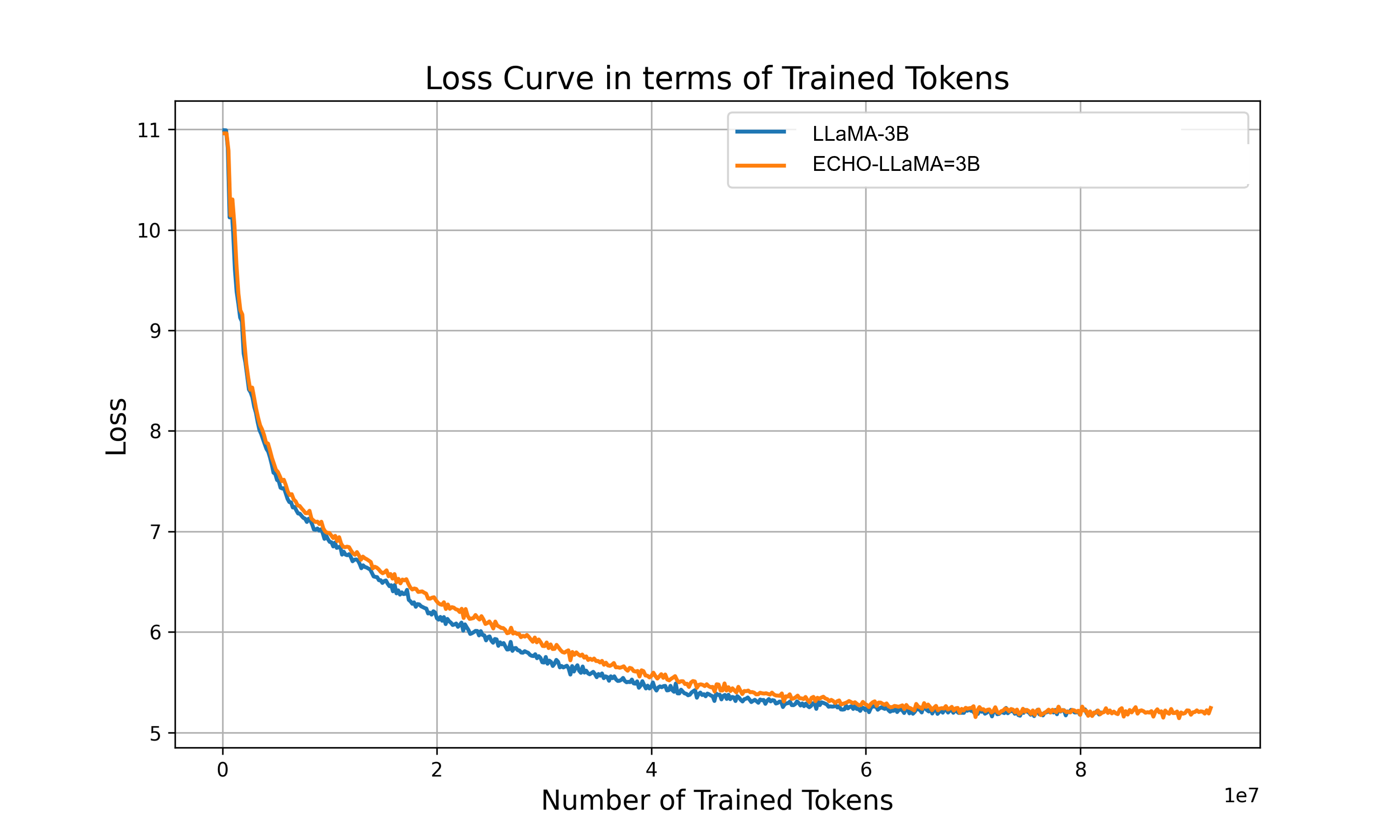} \\
        \multicolumn{3}{c}{\textbf{NPU (910B) Results}}
    \end{tabular}
    \caption{Training results for LLaMA-3B on GPU (V100) and NPU (910B). The columns represent (1) Train Loss vs. Steps, (2) Train Loss vs. Train Time, and (3) Train Loss vs. Number of Train Tokens.}
    \label{fig:3B_results}
\end{figure*}

\begin{figure*}[ht]
    \centering
    \subsection*{LLaMA-7B Model Training Results}
    \begin{tabular}{ccc}
        \includegraphics[width=0.3\textwidth]{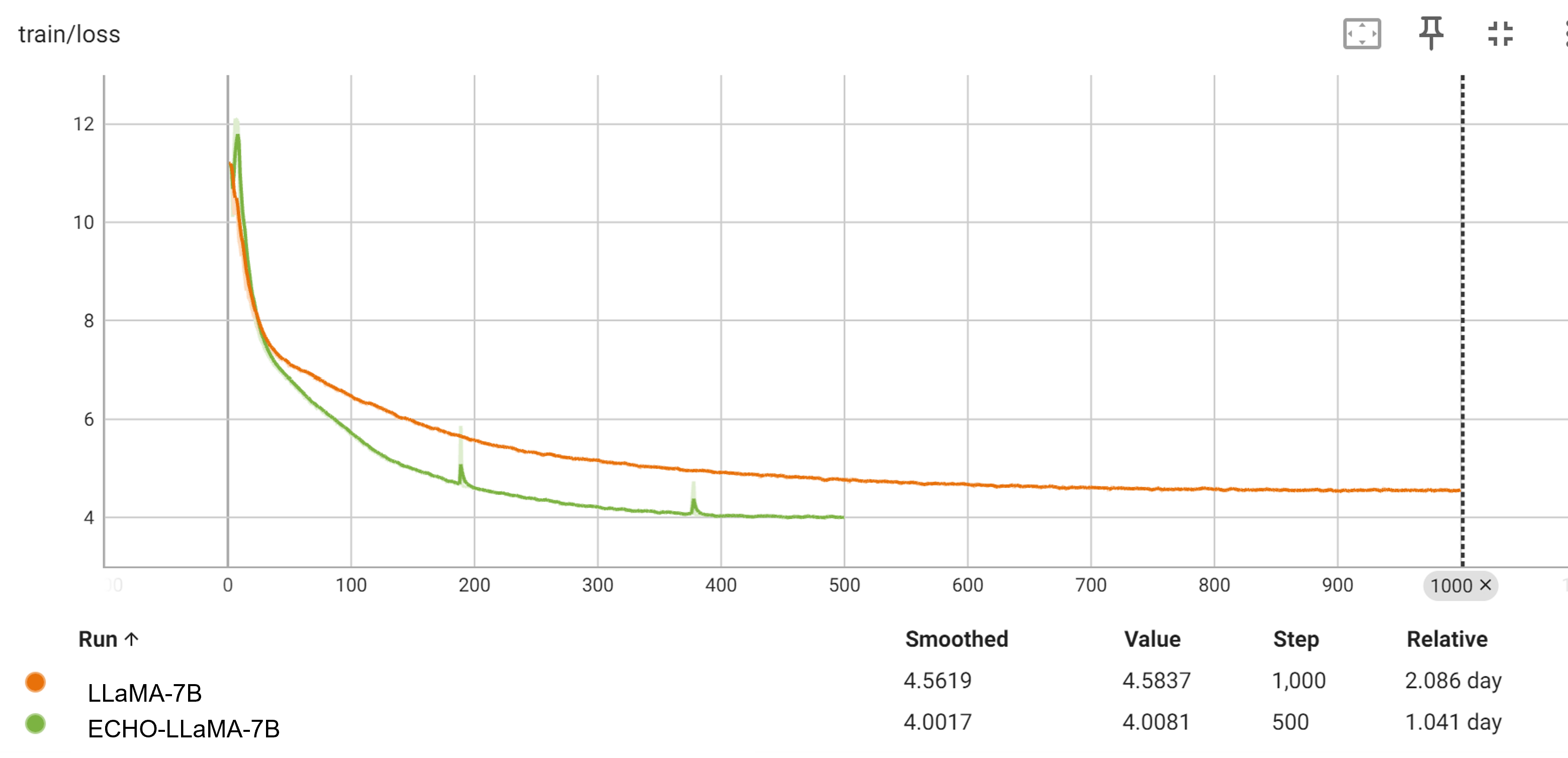} &
        \includegraphics[width=0.3\textwidth]{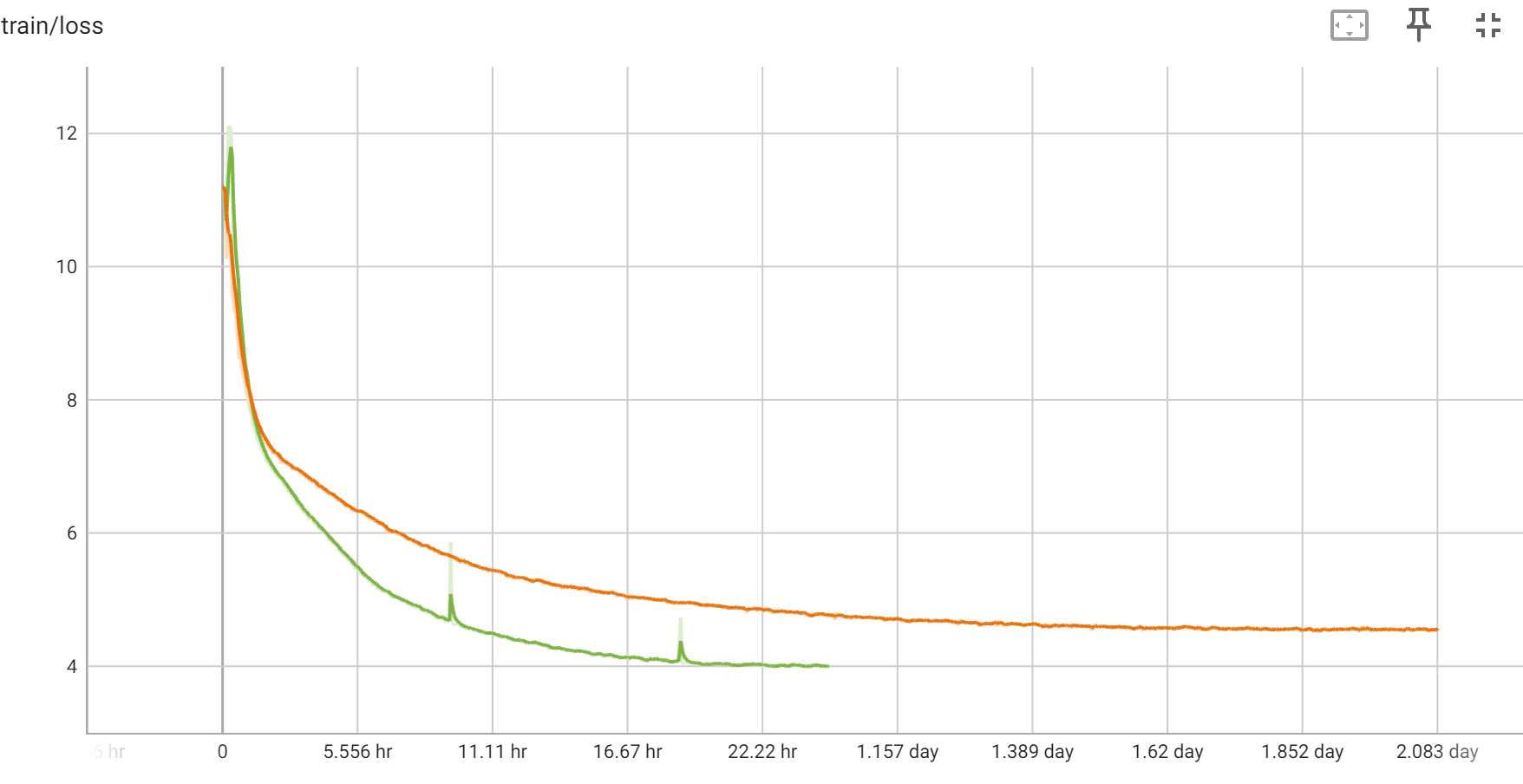} &
        \includegraphics[width=0.3\textwidth]{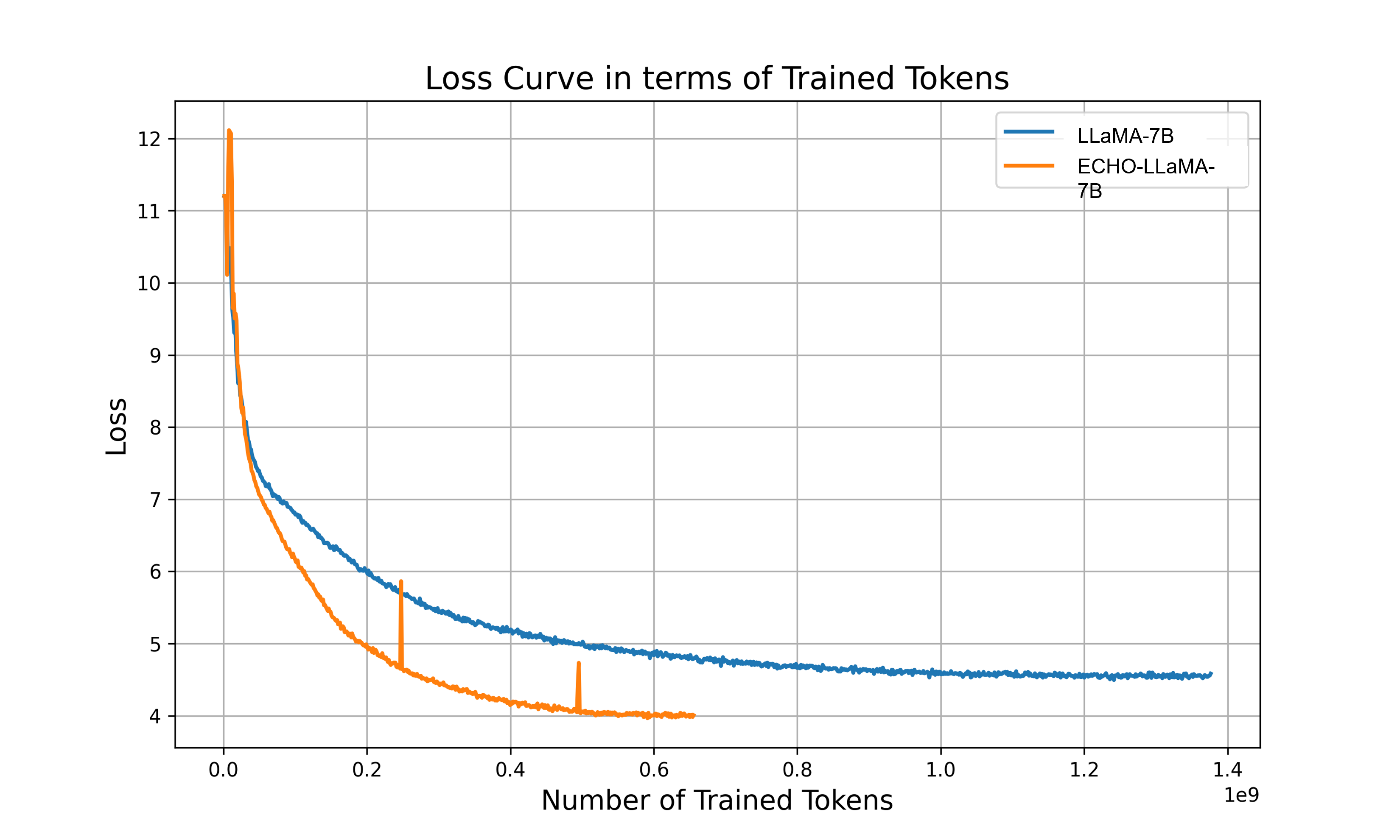} \\
        \multicolumn{3}{c}{\textbf{GPU (V100) Results}} \\
        \includegraphics[width=0.3\textwidth]{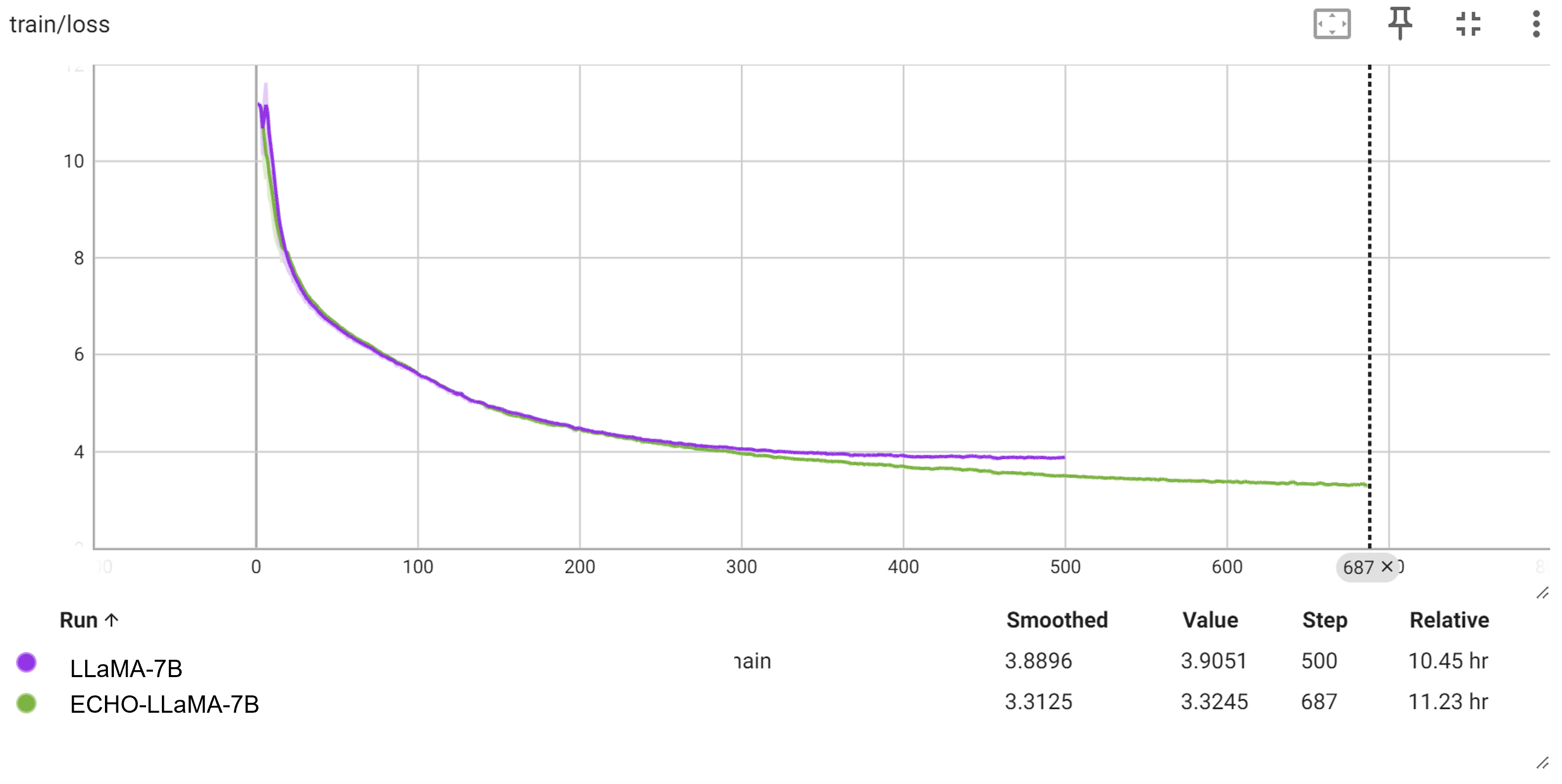} &
        \includegraphics[width=0.3\textwidth]{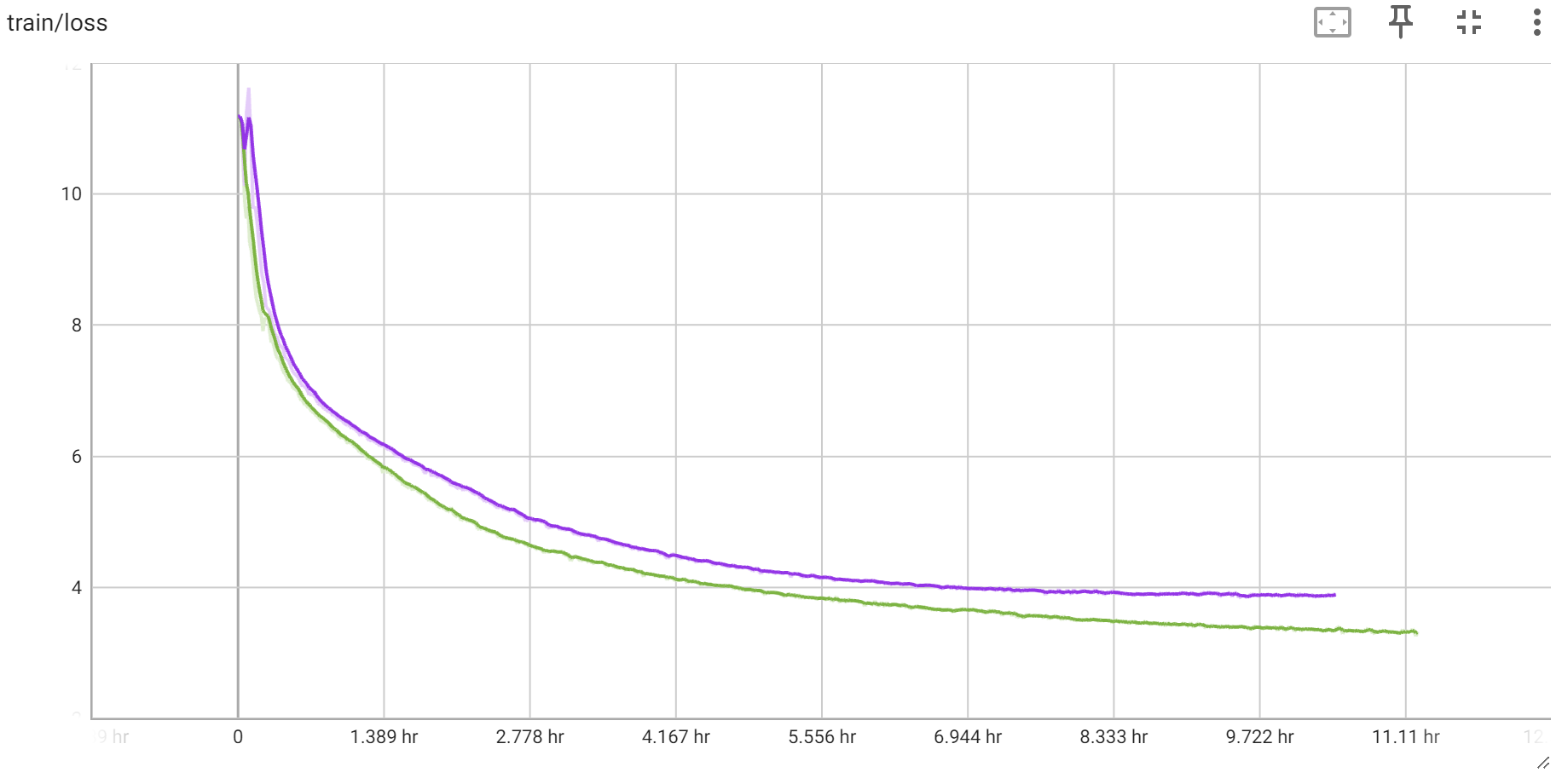} &
        \includegraphics[width=0.3\textwidth]{Figs/llama-125/npu-910b/llama-125-token-loss-npu910b.png} \\
        \multicolumn{3}{c}{\textbf{NPU (910B) Results}}
    \end{tabular}
    \caption{Training results for LLaMA-7B on GPU (V100) and NPU (910B). The columns represent (1) Train Loss vs. Steps, (2) Train Loss vs. Train Time, and (3) Train Loss vs. Number of Train Tokens.}
    \label{fig:7B_results}
\end{figure*}

To compare GPU and NPU performance across different LLM sizes, we organize the results in a structured figure layout. Figures \ref{fig:125M_results}, \ref{fig:1.1B_results}, \ref{fig:3B_results}, and \ref{fig:7B_results} present the results, where Each model size has a separate figure, and each figure contains two rows:
\begin{itemize}
    \item The \textbf{first row} contains plots for GPU (V100).
    \item The \textbf{second row} contains plots for NPU (910B).
\end{itemize}
Within each row, there are three subfigures, corresponding to the three types of plots. 
The left subfigure illustrate the training loss behavior in terms of training steps, the middle subfigure describes the training loss in terms of time, and the right one shows the same loss in terms of number of training tokens. As the plots depicts, the ECHO versions suppress or compete with the baselines, particularly in 7B model size. This observation shows that by increasing the size of model, the impact of ECHO-LLaMA becomes more strong. This is because in larger model sizes, the memory and compute demands grow significantly, making efficient KV sharing crucial for reducing memory overhead and improving throughput.  

\section{Comparison of Incremental vs. Full-Stage Shared-KV Fine-Tuning}
\label{appendix:FS-vs-Inc}
To evaluate the effectiveness of incremental strategy for applying shared key-value (KV) representations, we compare \textbf{Incremental Sharing}, where shared-KV layers are introduced gradually during training starting from the final layer, and \textbf{Full-Stage Sharing}, where a fixed subset of layers (25\% or 50\%) are converted to shared-KV at once and jointly fine-tuned.

\begin{figure*}[ht]
    \centering
    \includegraphics[width=\textwidth]{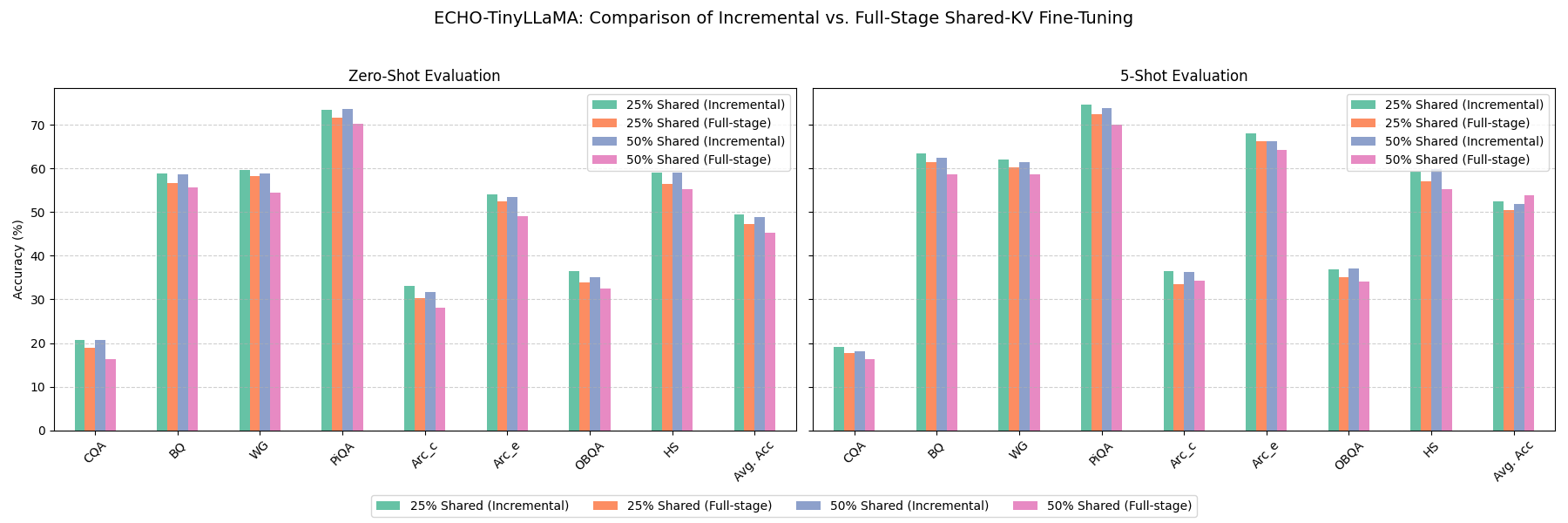}
    \caption{\textbf{LLaMA-7B}: Comparing Incremental Sharing over Full-Stage Sharing for shared-KV adaptation.}
    \label{fig:five_shot}
\end{figure*}

\begin{figure*}[t]
    \centering
    \includegraphics[width=\textwidth]{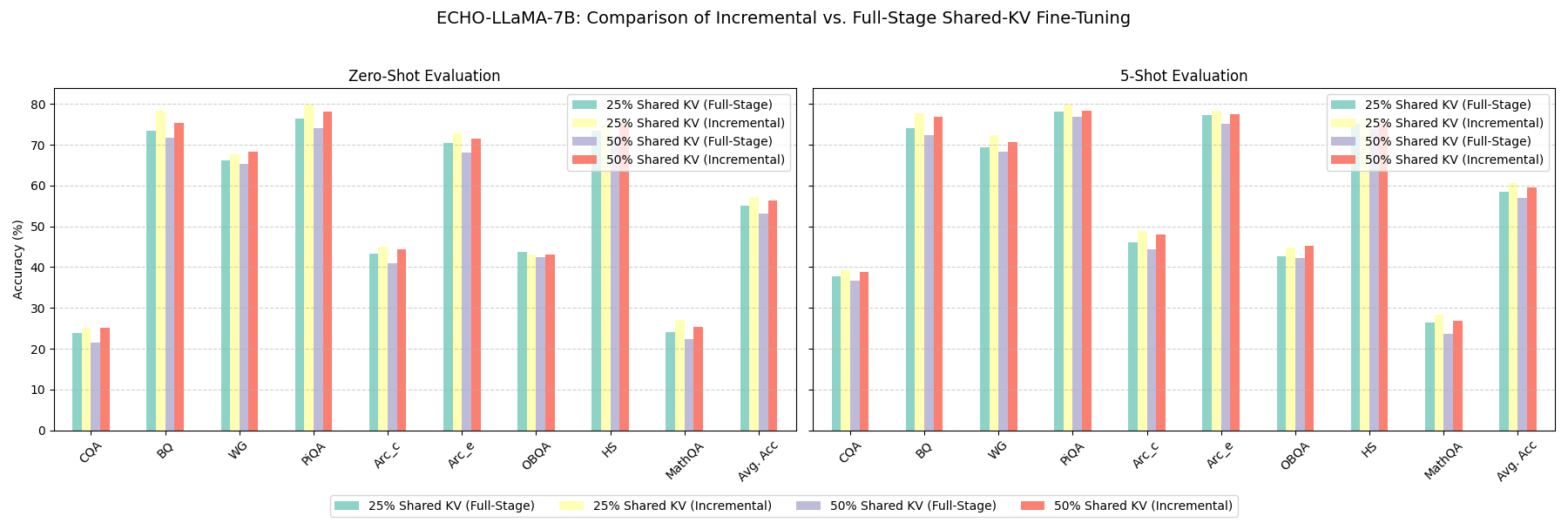}
    \caption{\textbf{LLaMA-7B}: Comparing Incremental Sharing over Full-Stage Sharing for shared-KV adaptation.}
    \label{Zero-ECHO-LLama-7B-Inc}
\end{figure*}

Figures \ref{fig:five_shot} and \ref{Zero-ECHO-LLama-7B-Inc} present the results for zero-shot and 5-shot evaluations across diverse benchmark datasets. Incremental Sharing consistently outperforms Full-Stage Sharing at both 25\% and 50\% sharing ratios, indicating that incremental adaptation enables the model to better preserve pre-trained knowledge while integrating structural modifications. This improvement is most notable in datasets such as Arc-c and OBQA, where the performance gap exceeds 2\%.

\section{More Evaluation on ECHO Archtecture}
\label{more-exp}
To further validate the effectiveness and generalizability of our proposed ECHO mechanism, we applied it to multiple publicly available models with LLaMA-style architectures. Table ~\ref{tab:selected_models_eval} presents a detailed comparison of baseline models and their ECHO-enhanced counterparts across a broad range of standard evaluation benchmarks, including MMLU~\cite{hendryckstest2021,hendrycks2021ethics}, C-Eval~\cite{huang2023ceval}, GSM8K~\cite{cobbe2021gsm8k}, MATH~\cite{hendrycks2021measuring}, HumanEval~\cite{chen2021evaluating}, and MBPP~\cite{austin2021program,touvron2023llama}.

We examine LLaMA2-7B~\cite{touvron2023llama}. The results show that applying ECHO with 25\% shared cross-decoder layers leads to a consistent improvement in average performance (22.56 vs. 22.15), with minor gains across multiple tasks. However, pushing to 50\% shared layers introduces degradation (average drops to 21.26), suggesting that over-sharing can hurt task-specific expressivity at this scale.

Given that the Qwen models~\cite{qwen} adopt a transformer architecture closely aligned with LLaMA—including the use of rotary positional embeddings and RMSNorm—we extended the ECHO framework to Qwen-1.8B and Qwen-7B as well. For Qwen-1.8B, applying ECHO with 25\% shared decoders marginally improves the average score (27.8 vs. 27.5), while the 50\% configuration results in performance decline. Similar trends hold for Qwen-7B, where the 25\% variant achieves the best overall average (41.31), modestly surpassing the baseline, while the 50\% configuration again incurs noticeable drop-offs.

These results collectively reinforce two key findings: (1) ECHO reliably improves efficiency without sacrificing performance when applied conservatively (e.g., 25\% sharing), and (2) the technique is transferable across architectures that share foundational transformer principles, such as LLaMA and Qwen.

\begin{table*}[ht]
\centering
\renewcommand{\arraystretch}{1.2}
\resizebox{\textwidth}{!}{%
\begin{tabular}{|l|c|c|c|c|c|c|c|}
\hline
\rowcolor{gray!30}
\textbf{Model} & \textbf{MMLU (5-shot)} & \textbf{C-Eval (5-shot)} & \textbf{GSM8K (8-shot)} & \textbf{MATH (4-shot)} & \textbf{HumanEval (0-shot)} & \textbf{MBPP (3-shot)} & \textbf{Avg.} \\ \hline
LLaMA2-7B (baseline)       & 46.8 & 32.5 & 16.7 & 3.3  & 12.8 & 20.8  & 22.15 \\ \hline
ECHO-LLaMA2-7B-25\%       & 46.9 & 33.4 & 17.2 & 3.6  & 13.5  & 20.8 & 22.56 \\ \hline
ECHO-LLaMA2-7B-50\%  & 46.2 & 31.8 & 15.6 & 2.1  & 12.2 & 19.7  & 21.26 \\ \hline \hline
Qwen-1.8B (baseline) & 45.3 & 56.1 & 32.3 & 2.3  & 15.2 & 14.2  & 27.5 \\ \hline
ECHO-Qwen-1.8B-25\% & 45.8 & 56.3 & 32.6 & 2.4  & 15.5 & 14.7 & 27.8 \\ \hline
ECHO-Qwen-1.8B-50\% & 44.1 & 54.7 & 30.8 & 1.8  & 13.9 & 12.7 & 26.33 \\ \hline \hline
Qwen-7B (baseline)   & 58.2 & 63.5 & 51.7 & 11.6 & 29.9 & 31.6  & 41.08 \\ \hline
ECHO-Qwen-7B-25\%   & 58.4 & 63.8 & 52.4 & 12.1 & 29.7 & 31.5  & 41.31 \\ \hline
ECHO-Qwen-7B-50\%   & 57.3 & 62.2 & 50.4 & 10.4 & 28.6 & 29.8 & 39.78 \\ \hline
\end{tabular}%
}
\caption{Evaluating the generalizability of the ECHO mechanism across LLaMA and Qwen architectures on diverse reasoning and coding benchmark. The results are reported in terms of accuracy.}
\label{tab:selected_models_eval}
\end{table*}

\begin{table*}[ht]
\centering
\renewcommand{\arraystretch}{1.2}
\resizebox{\textwidth}{!}{%
\begin{tabular}{|l|c|c|c|c|c|}
\hline
\rowcolor{gray!30}
\textbf{Model} & \textbf{MMLU (5-shot)} & \textbf{AGIEval-En (5-shot)} & \textbf{Arc-c (25-shot)} & \textbf{SQuAD (1-shot)} & \textbf{Avg.} \\ \hline
LLaMA3.2-1B (baseline)       &32.1  &23.1  &32.7 &49.0  &34.22 \\ \hline
ECHO-LLaMA3.2-1B-25\%        &32.4  &23.3  &32.6  &49.3   &34.40 \\ \hline
ECHO-LLaMA3.2-1B-50\%        &31.3  &21.5  &30.4  &47.8   &32.75 \\ \hline \hline
LLaMA3.2-3B (baseline)       &58.0 &39.1  &69.1  &67.7  &58.47 \\ \hline
ECHO-LLaMA3.2-3B-25\%        &58.3  &39.2 &69.0 &67.6   &58.52 \\ \hline
ECHO-LLaMA3.2-3B-50\%        &56.7 &37.3 &68.1 &65.7 &56.95 \\ \hline \hline
LLaMA3.2-8B (baseline)       &66.7 &47.7 &79.6 &69.7 &65.92 \\ \hline
ECHO-LLaMA3.2-8B-25\%        &66.8 &47.9 &79.5 &69.9 & 66.02\\ \hline
ECHO-LLaMA3.2-8B-50\%        &65.4 &47.1 &78.4 &68.4 &64.82 \\ \hline

\end{tabular}
}
\caption{Evaluating the generalizability of the ECHO mechanism across several LLaMA3.2 models  on diverse benchmark. The results are reported in terms of accuracy.}
\label{tab:selected_models_eval-2}
\end{table*}

\section{MFU-Loss-Speed}
Figure \ref{fig:npu_tokens_loss} compares the training throughput (Tokens/sec), Model FLOPs Utilization (MFU), and final loss across various LLaMA models and their ECHO-LLaMA counterparts. The ECHO-LLaMA versions consistently demonstrate improvements in MFU for configurations with 1, 4, and 8 devices, while achieving lower or comparable final loss compared to their baselines. Notably, ECHO-LLaMA models, such as ECHO-LLaMA-125M and ECHO-TinyLLaMA, exhibit significant speed improvements, as indicated by the purple bars, while maintaining competitive or better loss values. These results explains the effectiveness of ECHO-LLaMA's architecture in enhancing both training efficiency and model performance.

\section{Questions / Answers}
This section addresses several questions relevant to the research.

\subsection{Q1. Why is there no direct comparison with YOCO models?}
First, no official checkpoints for YOCO models have been released publicly, and a full pretraining of YOCO models from scratch is beyond the scope of this work. Second, while YOCO and ECHO-LLAMA both adopt a shared-KV mechanism, their methodologies differ fundamentally. YOCO requires a heavy pretraining phase from scratch, whereas ECHO-LLAMA proposes a lightweight fine-tuning approach that transforms existing pre-trained models into ECHO architectures. For these reasons, we did not include a direct comparison with YOCO in our evaluation.

\subsection{Q2. Why does the training throughput speedup degrade significantly beyond a certain model size threshold?
}
The reduced speedup observed for models exceeding a certain size threshold is primarily due to the increasing dominance of non-KV components—such as MLP layers—in the overall compute cost. As the model size grows, the proportion of parameters and computation attributed to these non-shared components becomes significantly larger than that of the shared KV layers. Consequently, the relative benefit of KV-sharing diminishes, leading to a lower overall throughput speedup.

\section{Model Configurations for LLaMA-125M and LLaMA-3B}
We provide detailed configuration settings for the LLaMA-125M and LLaMA-3B models, which were developed solely for this research. These configurations are designed by following the architectural patterns of TinyLLaMA, LLaMA-7B, and other higher versions of the LLaMA family. We adhered to the scaling strategy used in LLaMA models, ensuring proportional ratios for hidden size, intermediate size, number of layers, and attention heads as the models scale up.

\subsection{LLaMA-125M Configuration}
The following configuration outlines the settings for the LLaMA-125M model:
\small
\begin{verbatim}
{
  "architectures": ["ECHOLLaMAForCausalLM"],
  "hidden_size": 768,
  "intermediate_size": 2048,
  "num_hidden_layers": 12,
  "num_attention_heads": 12,
  "max_position_embeddings": 2048,
  "vocab_size": 32000,
  "rotary_emb_base": 10000,
  "tie_word_embeddings": False,
  "use_cache": True,
  "layer_norm_epsilon": 1e-5,
  "init_std": 0.02,
  "torch_dtype": "float16",
  "model_type": "echo_llama",
  "pad_token_id": None,
  "bos_token_id": 1,
  "eos_token_id": 2
}
\end{verbatim}

\subsection{LLaMA-3B Configuration}
The following configuration outlines the settings for the LLaMA-3B model:
\small
\begin{verbatim}
{
  "architectures": ["ECHOLLaMAForCausalLM"],
  "hidden_size": 3072,
  "intermediate_size": 8192,
  "num_hidden_layers": 26,
  "num_attention_heads": 24,
  "max_position_embeddings": 2048,
  "vocab_size": 32000,
  "rotary_emb_base": 10000,
  "tie_word_embeddings": False,
  "use_cache": True,
  "layer_norm_epsilon": 1e-5,
  "init_std": 0.02,
  "torch_dtype": "float16",
  "model_type": "echo_llama",
  "pad_token_id": None,
  "bos_token_id": 1,
  "eos_token_id": 2
}
\end{verbatim}

\end{document}